\documentclass[10pt,twocolumn,letterpaper]{article}

\usepackage{cvpr}              

\definecolor{cvprblue}{rgb}{0.21,0.49,0.74}
\usepackage[pagebackref,breaklinks,colorlinks,allcolors=cvprblue]{hyperref}

\usepackage{multirow, booktabs}
\usepackage{comment}
\usepackage{colortbl} 
\usepackage{xcolor}   
\usepackage{caption,setspace}
\usepackage{subcaption}

\newcommand{\ul}[1]{\underline{#1}}

\def\paperID{14} 
\def\confName{CVPR}
\def\confYear{2025}

\title{XYScanNet: A State Space Model for Single Image Deblurring}

\author{
    Hanzhou Liu$^{1}$ \quad Chengkai Liu$^{1}$ \quad Jiacong Xu$^{2}$ \quad Peng Jiang$^{1}$ \quad Mi Lu$^{1}$ \\
    $^1$Texas A\&M University \quad $^2$Johns Hopkins University\\
    {\tt\small \{hanzhou1996, liuchengkai, maskjp, milu0722\}@tamu.edu, jxu155@jhu.edu}\\
    \small{\url{https://github.com/HanzhouLiu/XYScanNet}}
}

\begin{document}
\maketitle
\begin{abstract}
Deep state-space models (SSMs), like recent Mamba architectures, are emerging as a promising alternative to CNN and Transformer networks.
Existing Mamba-based restoration methods process visual data by leveraging a flatten-and-scan strategy that converts image patches into a 1D sequence before scanning. 
However, this scanning paradigm ignores local pixel dependencies and introduces spatial misalignment by positioning distant pixels incorrectly adjacent, which reduces local noise-awareness and degrades image sharpness in low-level vision tasks. 
To overcome these issues, we propose a novel slice-and-scan strategy that alternates scanning along intra- and inter-slices. We further design a new Vision State Space Module (VSSM) for image deblurring, and tackle the inefficiency challenges of the current Mamba-based vision module. Building upon this, we develop XYScanNet, an SSM architecture integrated with a lightweight feature fusion module for enhanced image deblurring. XYScanNet, maintains competitive distortion metrics and significantly improves perceptual performance. Experimental results show that XYScanNet enhances KID by $17\%$ compared to the nearest competitor.
\end{abstract}    
\section{Introduction}
\label{sec:intro}
Single-image deblurring aims to restore a sharp image from a blurred one, typically framed as an inverse filtering problem~\cite{fergus2006removing,pan2016blind,yan2017image,chen2019blind,zhang2022deep}.
With the rapid development of deep learning, Convolutional Neural Networks (CNNs)~\cite{nah2017deep, kupyn2018deblurgan, kupyn2019deblurgan, zhang2019deep, suin2020spatially, li2021perceptual, li2021advanced, cho2021rethinking, zamir2021multi, liu2023real} have become the dominant approach for image deblurring. Recently, Transformer-based models \cite{zamir2022restormer, wang2022uformer, tsai2022stripformer, kong2023efficient, mao2024loformer}  have also shown strong performance, leveraging their attention-based mechanisms.

\begin{figure}[tb]
  \centering
  \includegraphics[width=0.464\textwidth, trim = 2.8cm 2.8cm 2.8cm 2.8cm]{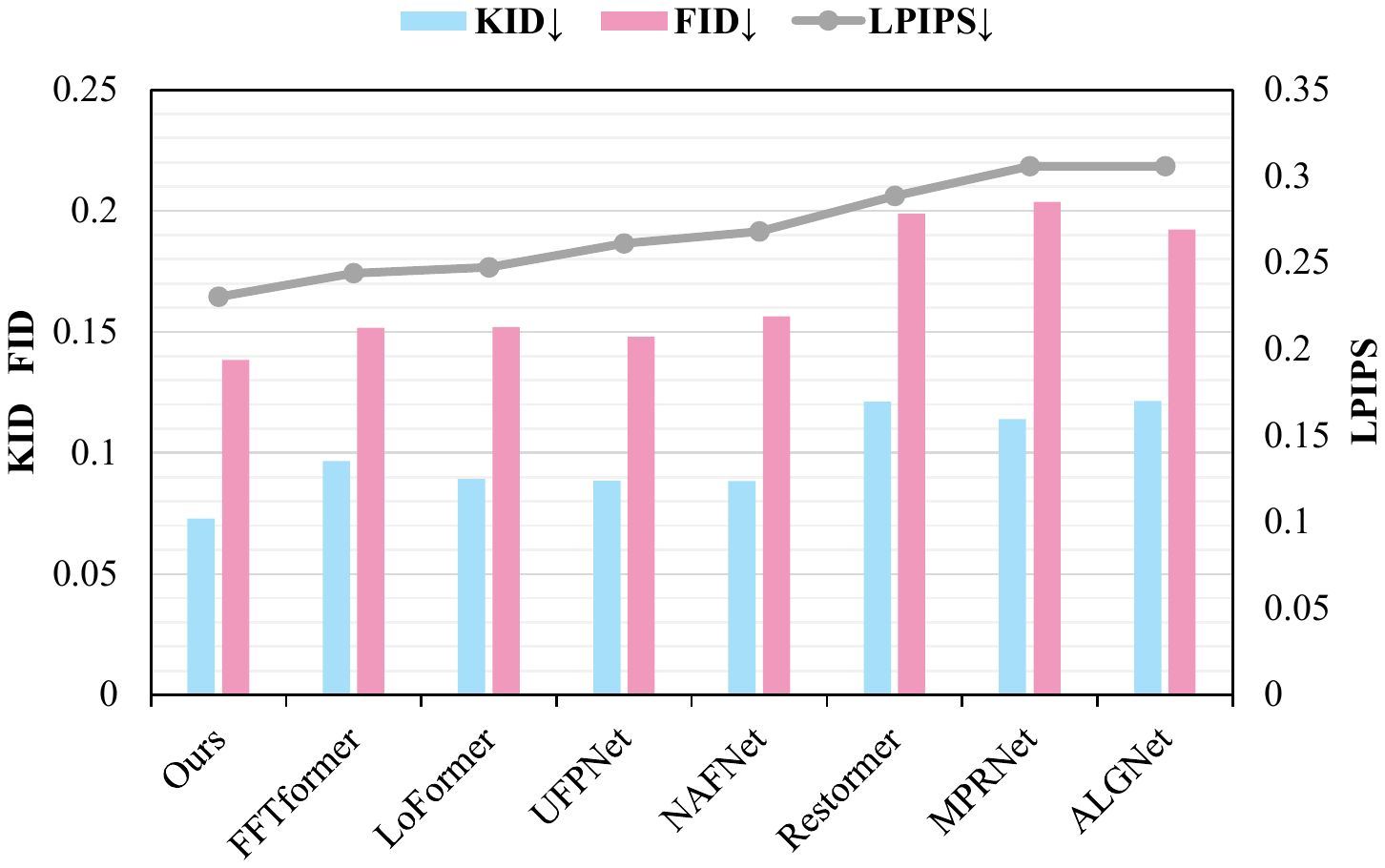}

  \caption{XYScanNet achieves state-of-the-art performance on the GoPro dataset~\cite{nah2017deep}, measured by the perceptual metrics normalized to $[0,1]$ for clear visualizations.}\label{figure:chart}
\end{figure}

\definecolor{tiffanyblue}{rgb}{0.04, 0.73, 0.71}
\begin{figure*}[tb]
  \centering
  \includegraphics[width=0.99\textwidth, trim = 0.cm 8.1cm 0cm 0.6cm]{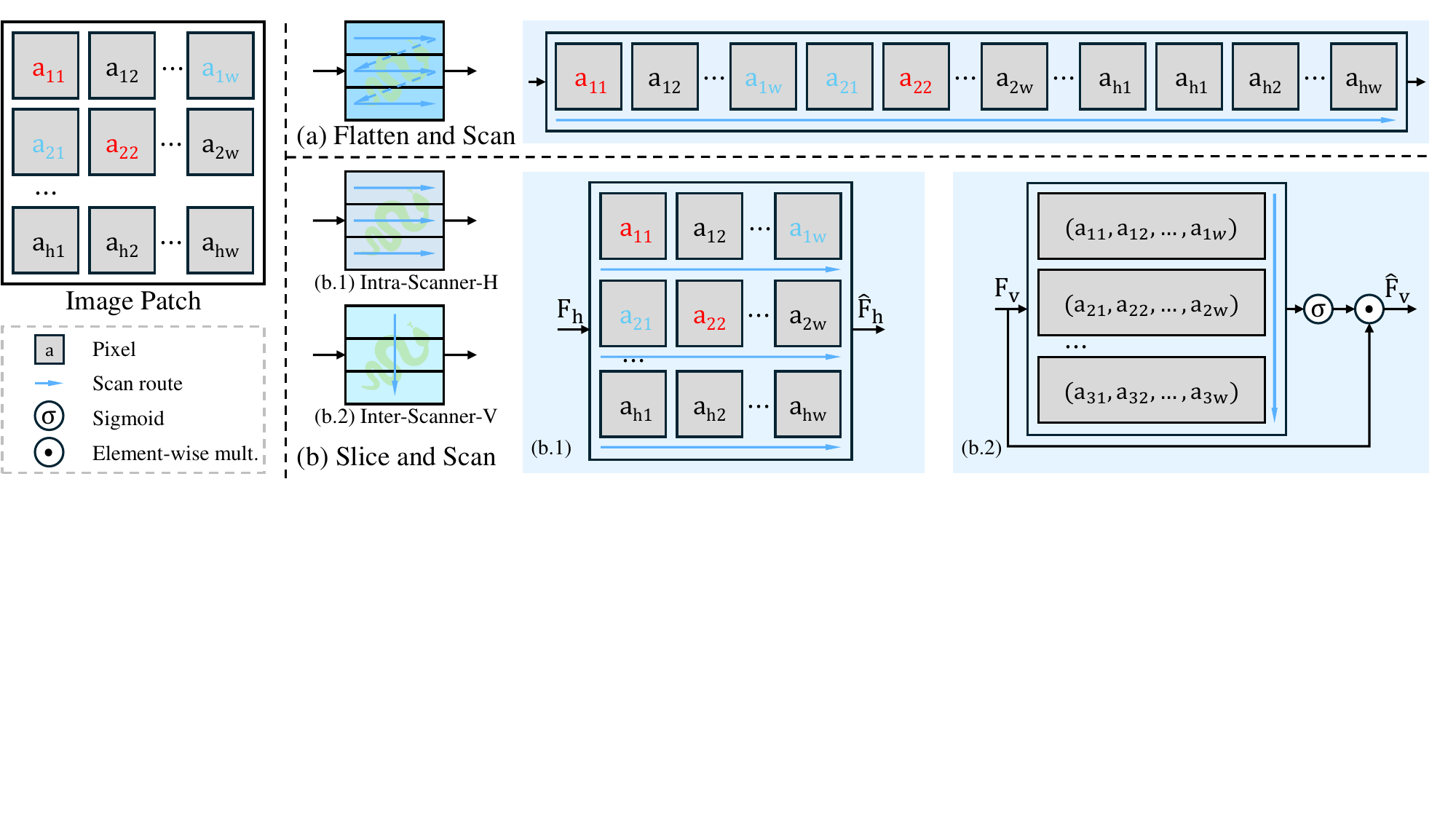}

  \caption{(a) The two issues of flatten-and-scan strategies in a single scanning route: adjacent \textcolor{red}{$a_{11}$} and \textcolor{red}{$a_{22}$} become distant ({\bf local pixel forgetting}), while distant \textcolor{tiffanyblue}{$a_{1w}$} and \textcolor{tiffanyblue}{$a_{21}$} are placed adjacent ({\bf spatial misalignment}). (b.1) Intra-Scanner in the horizontal branch; (b.2) Inter-Scanner in the vertical branch. Intra-Scanner-V and Inter-Scanner-H are symmetrically structured.
  }\label{figure:scan}
\end{figure*}

The recent Mamba architectures~\cite{gu2023mamba} combining state-space models (SSMs) with Selective Scan, are emerging as a promising alternative to CNNs and Transformers for vision tasks~\cite{zhu2024vision, liu2024vmamba}. However,~\cref{figure:scan}a shows that the existing flatten-and-scan strategies to process visual data for Mamba algorithms, resulting in two key challenges in a single scanning route, which degrade image sharpness on the image deblurring task. First, short-range dependencies are lost due to the flattening of image patches into a 1D sequence, known as {\bf local pixel forgetting}~\cite{guo2024mambair}, which can be easily resolved by integrating depth-wise convolutions. Second, distant pixels are incorrectly placed next to each other, disrupting spatial context, termed as {\bf spatial misalignment}. While four-way directional scanning in MambaIR~\cite{guo2024mambair} may implicitly mitigate the negative effects of {\bf spatial misalignment}, the brute-force quadrupling of scanning routes leads to a significant increase in computational cost.

To address the aforementioned concern, we introduce a straightforward slice-and-scan strategy that preserves well-aligned spatial relationships and solves {\bf spatial misalignment} efficiently.~\cref{figure:scan}b illustrates that our design consists of interleaved Intra-Scanners and Inter-Scanners, each made up of horizontal and vertical branches. The cross-directional design is inspired by previous studies~\cite{sun2015learning, tsai2022stripformer}, which maps blur motions onto the horizontal and vertical axes of the Cartesian coordinate system to predict the motion blur field in a blurred image. Intra-Scanners maintain pixel-level clarity for local blur estimation, while avoiding {\bf spatial misalignment} by preventing direct interactions between misaligned pixels. However, the network relying solely on Intra-Scanners can be computationally expensive and fails to capture cross-slice dependencies. To overcome these, we design lightweight Inter-Scanners to capture the global blur patterns at the slice level. We adopt an interleaved architecture which replaces half of the Intra-Scanners with Inter-Scanners. By integrating these intuitive while effective ideas, we propose a novel Vision State Space Module (VSSM) for single image deblurring, offering significant improvements in both efficiency and effectiveness over the existing VSSM in MambaIR~\cite{guo2024mambair}.

Besides, we introduce a lightweight feature fusion module to efficiently leverage multi-level features, using only half the parameters of the AFF~\cite{cho2021rethinking}. Combining these advancements, we present XYScanNet, a novel SSM for image deblurring that achieves competitive distortion metrics while markedly enhancing perceptual performance.

The key contributions of this paper are three-fold:
\begin{itemize} 
\item We identify the long-standing issue of {\bf spatial misalignment} in Mamba-based vision work and address it with an efficient and straightforward slice-and-scan approach.
\item We develop a new vision state space module ({\bf VSSM}) that clearly reduces computational costs with improved visual fidelity over the existing Mamba-based counterpart~\cite{guo2024mambair}. 
\item We design a state space model for single image deblurring, achieving {\bf competitive distortion metrics and impressive perceptual performance} on multiple datasets.
\end{itemize}
\section{Related Work}
\label{sec:related_work}

\noindent
{\bf Single Image Deblurring.}
Over the past decade, CNN-based methods~\cite{nah2017deep, tao2018scale, zhang2019deep, li2021advanced, cho2021rethinking, liu2023real} have become the standard for image deblurring, outperforming traditional approaches~\cite{fergus2006removing, joshi2009image, zhang2013multi, ren2016image, pan2016blind, yan2017image, chen2019blind} by leveraging large-scale visual data to learn general priors~\cite{koh2021single,zhang2022deep,zamir2021multi}. Inspired by the success of Transformers in Natural Language Processing, recent work has adapted them for low-level computer vision tasks like deblurring~\cite{zamir2022restormer, wang2022uformer, tsai2022stripformer, kong2023efficient, liu2024deblurdinat}. For example, the frequency-based network FFTformer~\cite{kong2023efficient} achieves state-of-the-art PSNR and SSIM scores on mainstream datasets, but demonstrates limited perceptual metrics. In terms of perceptual performance, a diffusion-based approach~\cite{whang2022deblurring} delivers strong metric scores as rated by KID (Kernel Inception Distance)~\cite{binkowski2018demystifying}, FID (Fréchet Inception Distance)~\cite{heusel2017gans}, LPIPS~\cite{zhang2019deep} and NIQE~\cite{mittal2012making}, despite its lower distortion metrics. Our state-space model focuses on maintaining competitive distortion metrics and enhancing perceptual scores, distinguishing it from recent deblurring studies that struggle to excel in both aspects.


\begin{figure*}[tb]
  \centering
  \includegraphics[width=0.985\textwidth, trim = 0.8cm 6.1cm 5.4cm 0.5cm]{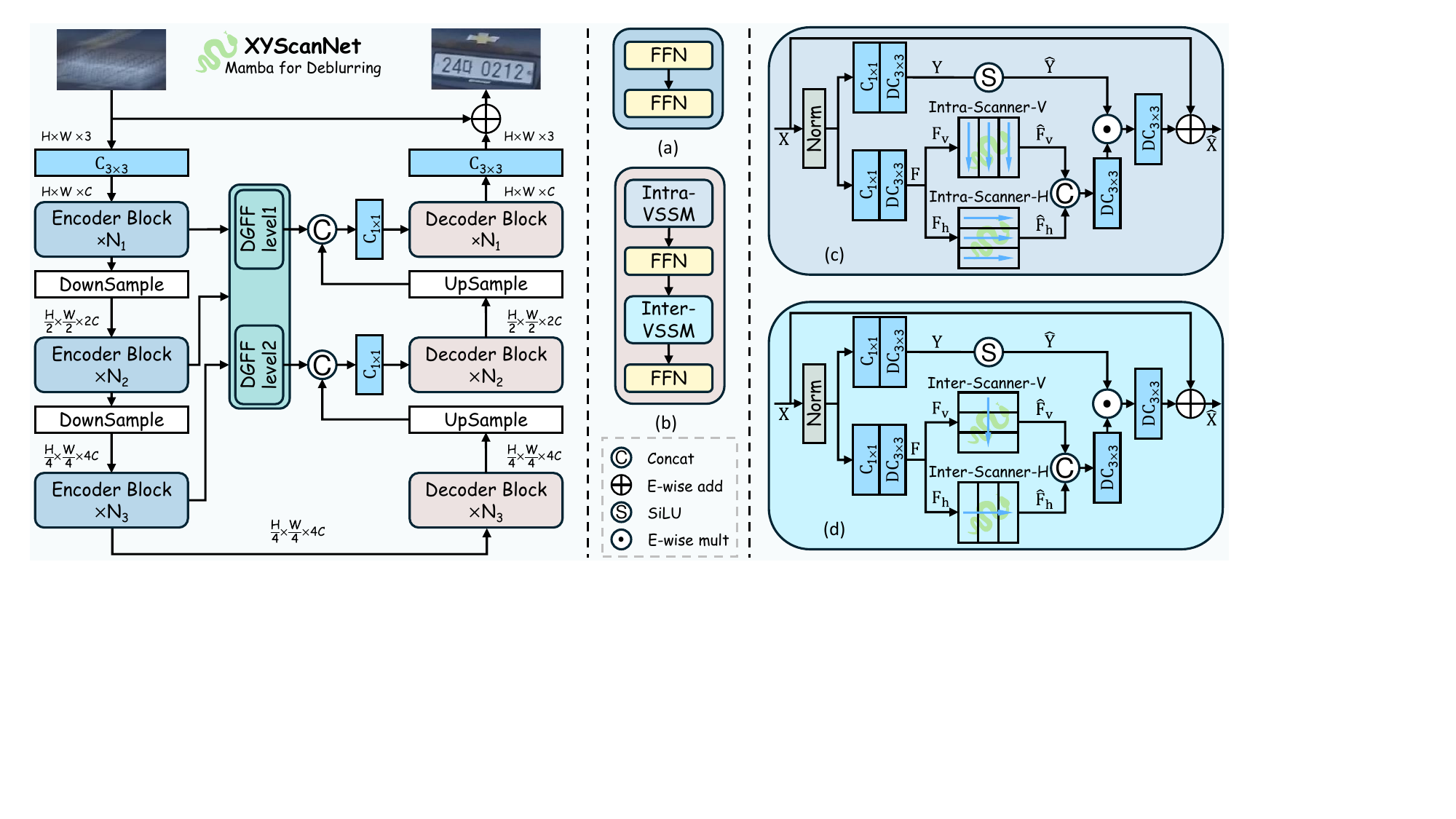}

  \caption{{\bf XYScanNet} is a three-level U-Net. (a) An encoder block. (b) A decoder block. (c) {\bf Intra-VSSM} incorporates $3\times3$ depth-wise convolutions to prevent local pixel forgetting~\cite{guo2024mambair} and the dual-branch {\bf Intra-Scanners} (refer to~\cref{figure:scan} b.1) to address spatial misalignment. (d) Similarly, {\bf Inter-VSSM} incorporates convolutions and the parallel {\bf Inter-Scanners} (see~\cref{figure:scan} b.2).
  }\label{figure:architecture}
\end{figure*}
\noindent
{\bf State Space Models.}
State-space models (SSMs) have recently seen promising adaptations in deep learning~\cite{gu2020hippo,gu2021combining,gu2021efficiently,fu2022hungry,smith2022simplified,mehta2022long,gu2023mamba,poli2023hyena,patro2024mamba}. The latest advancement, SSM with Selective {\bf Scan} (Mamba-S6)~\cite{gu2023mamba} enables selective focus on inputs and optimizes for hardware-aware parallelism, positioning it as a potential alternative to Transformers. 

Recent vision-based Mamba architectures have focused on enhancing contextual modeling for task-specific feature learning. Vision Mamba~\cite{zhu2024vision} introduces a bidirectional Mamba mechanism, while VMamba~\cite{liu2024vmamba} employs a 2D Selective Scan (SS2D) module with four-way scanning for improved visual context capture. MambaIR~\cite{guo2024mambair} extends SS2D with the Vision State Space Module (VSSM), designed for low-level computer vision but hindered by high computational costs on full-resolution images. To improve efficiency, ALGNet~\cite{gao2024aggregating} integrates a simplified Mamba branch for global feature extraction and a parallel convolutional branch for local blurred pattern recognition. While these Mamba-based approaches effectively address {\bf local pixel forgetting}, they remain limited in tackling {\bf spatial misalignment} due to flatten-and-scan strategies. Although MambaIR~\cite{guo2024mambair} may implicitly handle this issue through brute-force repeated scanning routes in four directions, it results in a significant increase in computational cost. In this work, we provide a comprehensive comparison between our proposed VSSM and its counterpart in MambaIR~\cite{guo2024mambair}, highlighting improvements in both spatial alignment and computational efficiency.


\section{Methodology}
\label{sec:method}
\subsection{Preliminaries}\label{sec:revisit_ssm}
{\bf Selective State Space Models} (Mamba-S6)~\cite{gu2023mamba} introduce time-varying parameters to replace the constant counterparts in S4~\cite{gu2021efficiently}. The time-dependent parameters results that the output can no longer be expressed in the form of a convolution formula. To this end, a hardware-aware Selective Scan technique is proposed.

\medskip
\noindent
{\bf MambaIR}~\cite{guo2024mambair} leverages the advantage of Mamba, ultra-long sequence memorization, to activate more pixels for image restoration. However, ALGNet~\cite{gao2024aggregating} reports the limitations of MambaIR on image deblurring tasks, and our experimental results in~\cref{tab:mambair} further reveal that its core component, Vision State Space Module (VSSM), is inefficient in both training and inference phases. 

\subsection{Slice and Scan}\label{sec:slice_scan}
In this section, we present the core components of XYScanNet, Intra-Scanner and Inter-Scanner, based on the straightforward slice-and-scan scheme. Theirs roles in motion estimation have been {\bf bolded} in the text.

\medskip
\noindent
{\bf Intra-Scanner.}
As shown in~\cref{figure:scan} (b.1), from an input tensor $\mathbf{F_h}\in\mathbb{R}^{\hat{B}\times\hat{H}\times\hat{W}\times\hat{\frac{C}{2}}}$, Intra-Scanner-H first slices feature maps along the height dimension and combines it with the batch dimension, yielding a 1D sequence $\mathbf{F^{\prime}_h}\in\mathbb{R}^{(\hat{B}\hat{H})\times\hat{W}\times\hat{\frac{C}{2}}}$. This intra-slicing mechanism preserves the original contextual information at each height, mitigating concerns about {\bf spatial misalignment} caused by existing flattening strategies~\cite{guo2024mambair}. Next, we employ the Mamba-S6 algorithm with Selective Scan~\cite{gu2023mamba}
, outputting feature maps $\mathbf{F^{\prime}_h}$. {\bf By scanning $\mathbf{F^{\prime}_h}$ in a uni-directional manner, the Intra-Scanner-H captures pixel-level dependencies and detects localized motion blur along the horizontal direction.} The Intra-Scanner-V, constructed symmetrically, estimates the vertical projection of motion blur. Overall, the outputs of Intra-Scanner-H and -V are computed by:
\begin{equation}
  \begin{aligned} \label{eq:intra-scanner}
  &\mathbf{\hat{F}_{h}} = \mathrm{S6}[\mathrm{Reshape}(\mathbf{\hat{F}_{h}}, (\hat{B}\times\hat{H}, \hat{W}, \hat{\frac{C}{2}}))]\\
  &\mathbf{\hat{F}_{v}} = \mathrm{S6}[\mathrm{Reshape}(\mathbf{\hat{F}_{v}}, (\hat{B}\times\hat{W}, \hat{H}, \hat{\frac{C}{2}}))], \\
  \end{aligned}
\end{equation}
where $\mathrm{S6}$ denotes the Mamba-S6 algorithm~\cite{gu2023mamba}, and $\mathrm{Reshape}$ is a tensor shape transformation.

\medskip
\noindent
{\bf Inter-Scanner.}
Solely using Intra-Scanners causes high computational costs due to their pixel-level granularity and miss cross-slice information, which is essential for adaptive blur estimation. Unlike previous intra-slicing methods~\cite{tsai2022stripformer}, ours introduces a compression factor $\delta$ to improve efficiency. We explain Inter-Scanner-V at first. As shown in~\cref{figure:scan}b.2, from an input tensor $\mathbf{F_v}\in\mathbb{R}^{\hat{B}\times\hat{H}\times\hat{W}\times\hat{\frac{C}{2}}}$, the Intra-Scanner-V first slices feature maps along the height dimension and compress the width dimension by a factor of $\delta$. For simplicity, we set $\delta$ to $\frac{1}{\hat{W}}$, so that the compression process can be easily implemented by a global average pooling (GAP) function, yielding a 1D sequence $\mathbf{F^{\prime}_v}\in\mathbb{R}^{\hat{B}\times\hat{H}\times\hat{\frac{C}{2}}}$. After that, the Mamba-S6 algorithm~\cite{gu2023mamba} is applied to $\mathbf{F^{\prime}_v}$, and the resulting features pass through a $\mathrm{Sigmoid}$ activation function and is multiplied by the input $\mathbf{F_v}$, generating the output $\mathbf{\hat{F}_{v}}$. {\bf By scanning $\mathbf{F^\prime_{v}}$ in a uni-directional manner, Inter-Scanner-V efficiently captures cross-slice dependencies along the vertical direction and estimates blur with larger magnitudes.} The horizontal Intra-Scanner is constructed symmetrically. Overall, the outputs of Inter-Scanner-V and -H are computed by~\cref{eq:inter-scanner}.
\begin{equation}
  \begin{aligned} \label{eq:inter-scanner}
  &\mathbf{\hat{F}_{v}} = \mathrm{\sigma}\{\mathrm{S6}[\mathrm{GAP}(\mathbf{F_v}, (\hat{B}, \hat{H}, \hat{\frac{C}{2}}))]\}\odot \mathbf{F_v}\\
  &\mathbf{\hat{F}_{h}} = \mathrm{\sigma}\{\mathrm{S6}[\mathrm{GAP}(\mathbf{F_h}, (\hat{B}, \hat{W}, \hat{\frac{C}{2}}))]\}\odot \mathbf{F_h}, \\
  \end{aligned}
\end{equation}
where $\mathrm{GAP}$ is the global average pooling function {\bf with no learnable parameters}; $\mathrm{S6}$ denotes the Mamba-S6 algorithm with Selective Scan~\cite{gu2023mamba}; and $\sigma$ represents the Sigmoid activation function.

\subsection{Vision State Space Module}\label{sec:VSSM}
The batch dimension $\hat{B}$ is omitted for simplicity. As shown in~\cref{figure:architecture} (c) and (d), from an input $\mathbf{X} \in \mathbb{R}^{\hat{H}\times\hat{W}\times\hat{C}}$, VSSM first applies a layer normalization and produces two projections $\mathbf{Y}$ and $\mathbf{F}$ by cascaded pixel-wise and depth-wise convolutions; where $\hat{H}\times\hat{W}$ represents the spatial dimension and $\hat{C}$ is the channel number. The feature maps $\mathbf{Y}$ pass through a SiLU activation layer~\cite{hendrycks2016gaussian,Ramachandran2017SwishAS,elfwing2018sigmoid}, resulting non-linear features $\mathbf{\hat{Y}}$. Meanwhile, the other features $\mathbf{F}$ is equally divided  along the channel dimension, yielding $\mathbf{F}_\mathbf{v}$ and $\mathbf{F}_\mathbf{h}$. 
Motivated by the dual-branch design in~\cite{tsai2022stripformer},
our Mamba-based VSSM contains two parallel components, Scanner-Vertical (V) and -Horizontal (H), to estimate motion blur at different angles. These two parallel Scanners produce ${\mathbf{\hat{F}_{v}}}$ and $\mathbf{\hat{F}_{h}}$ correspondingly, which are concatenated and then multiplied by $\mathbf{\hat{Y}}$ for controlled feature propagation~\cite{zamir2022restormer}. Finally, an element-wise multiplication and addition are performed to obtain the output. As shown in~\cref{figure:architecture}b, {\bf by stacking VSSMs with interleaved Intra- and Inter-Scanners, each of which uses a cross-directional design, our network is capable of estimating blur patterns with varying magnitudes and angles.} The overall process of our VSSM is defined in~\cref{eq:vssm}.
\begin{equation}
  \begin{aligned} \label{eq:vssm}
  &\mathbf{\hat{X}} = \mathrm{SiLU(\mathbf{Y})}\odot \mathrm{DualScanner(\mathbf{F})}+\mathbf{X}\\
  &(\mathbf{Y}, \mathbf{F}) = W_{d}W_{p}\mathbf{\mathcal{L}(X)},
  \end{aligned}
\end{equation}
where $W_{d}$ and $W_{p}$ represent the depth-wise and pixel-wise convolution; $\mathrm{DualScanner}$ denotes Intra- or Inter-Scanners with a dual-branch design, as discussed in~\cref{sec:slice_scan}.


\subsection{Cross Level Feature Fusion}
\begin{figure}[tb]
  \centering
  \includegraphics[width=0.47\textwidth, trim = 0.5cm 13cm 16.5cm 0.3cm]{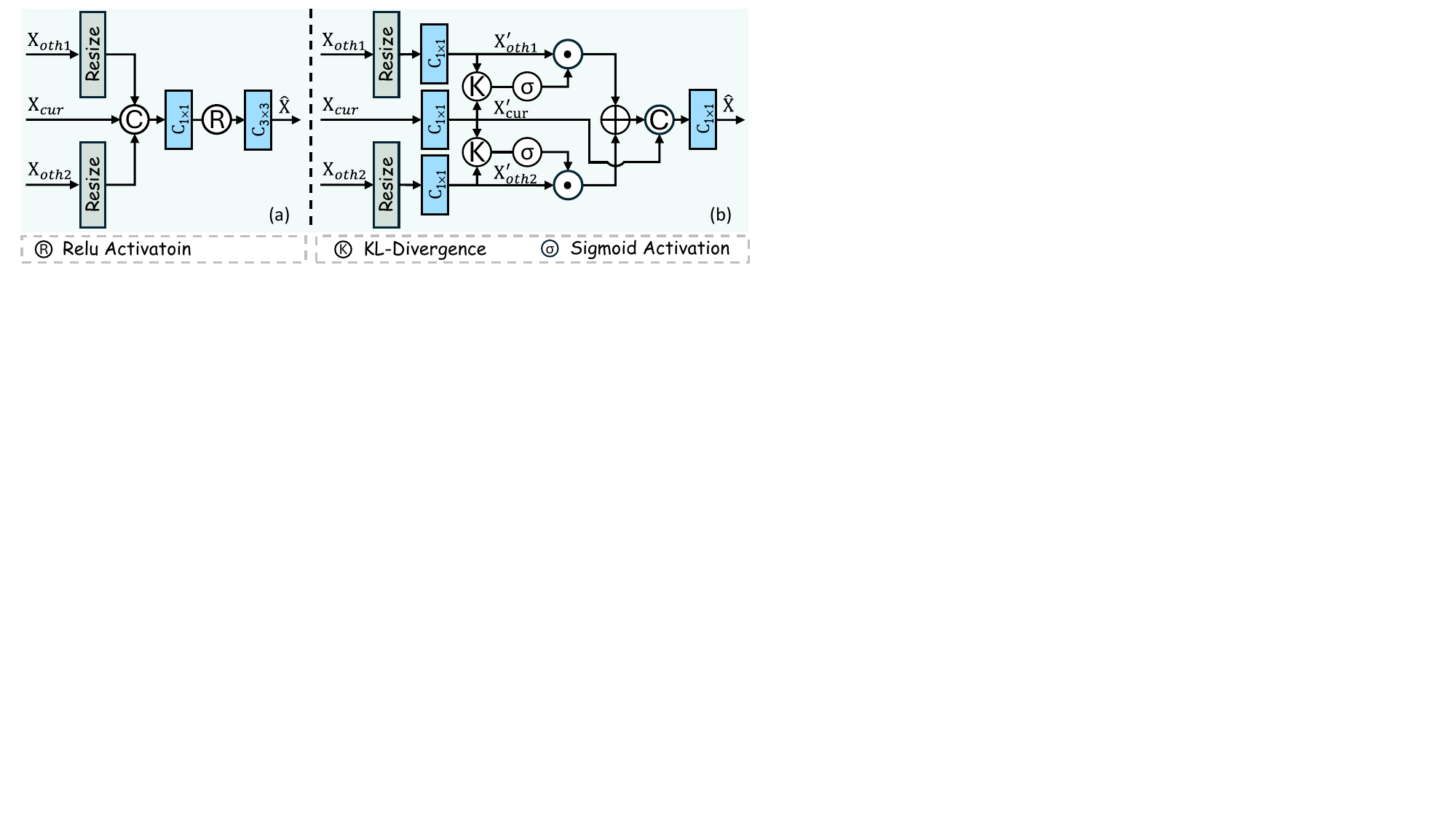}

  \caption{Structures of (a) asymmetric feature fusion (AFF)~\cite{cho2021rethinking}, and (b) dual gating feature fusion ({\bf DGFF}) modules.
  }\label{figure:ms}
\end{figure}

\begin{figure*}[tb]
  \captionsetup[subfigure]{labelformat=empty, justification=centering}
  \centering
  
  \begin{subfigure}[b]{0.11\textwidth}
    \centering
    \includegraphics[width=1\textwidth]{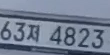}
    \caption{Reference}
  \end{subfigure}
  \hfill
  \vspace{0.1mm}
  \begin{subfigure}[b]{0.11\textwidth}
      \centering
      \includegraphics[width=1\textwidth]{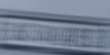}
      \caption{Blurred}
  \end{subfigure}
  \hfill
  \vspace{0.1mm}
  \begin{subfigure}[b]{0.11\textwidth}
      \centering
      \includegraphics[width=1\textwidth]{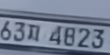}
      \caption{NAFNet}
  \end{subfigure}
  \hfill
  \vspace{0.1mm}
  \begin{subfigure}[b]{0.11\textwidth}
      \centering
      \includegraphics[width=1\textwidth]{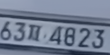}
      \caption{Restormer}
  \end{subfigure}
  \hfill
  \vspace{0.1mm}
  \begin{subfigure}[b]{0.11\textwidth}
      \centering
      \includegraphics[width=1\textwidth]{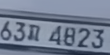}
      \caption{UFPNet}
  \end{subfigure}
  \hfill
  \vspace{0.1mm}
  \begin{subfigure}[b]{0.11\textwidth}
      \centering
      \includegraphics[width=1\textwidth]{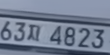}
      \caption{FFTformer}
  \end{subfigure}
  \hfill
  \vspace{0.1mm}
  \begin{subfigure}[b]{0.11\textwidth}
      \centering
      \includegraphics[width=1\textwidth]{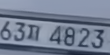}
      \caption{LoFormer}
  \end{subfigure}
  \hfill
  \vspace{0.1mm}
  \begin{subfigure}[b]{0.11\textwidth}
      \centering
      \includegraphics[width=1\textwidth]{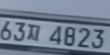}
      \caption{\bf Ours}
  \end{subfigure}
  \hfill
  \vspace{0.1mm}

  \begin{subfigure}[b]{0.11\textwidth}
    \centering
    \includegraphics[width=1\textwidth]{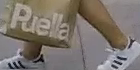}
    \caption{Reference}
  \end{subfigure}
  \hfill
  \vspace{0.1mm}
  \begin{subfigure}[b]{0.11\textwidth}
      \centering
      \includegraphics[width=1\textwidth]{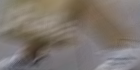}
      \caption{Blurred}
  \end{subfigure}
  \hfill
  \vspace{0.1mm}
  \begin{subfigure}[b]{0.11\textwidth}
      \centering
      \includegraphics[width=1\textwidth]{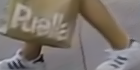}
      \caption{NAFNet}
  \end{subfigure}
  \hfill
  \vspace{0.1mm}
  \begin{subfigure}[b]{0.11\textwidth}
      \centering
      \includegraphics[width=1\textwidth]{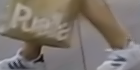}
      \caption{Restormer}
  \end{subfigure}
  \hfill
  \vspace{0.1mm}
  \begin{subfigure}[b]{0.11\textwidth}
      \centering
      \includegraphics[width=1\textwidth]{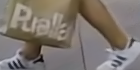}
      \caption{UFPNet}
  \end{subfigure}
  \hfill
  \vspace{0.1mm}
  \begin{subfigure}[b]{0.11\textwidth}
      \centering
      \includegraphics[width=1\textwidth]{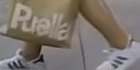}
      \caption{FFTformer}
  \end{subfigure}
  \hfill
  \vspace{0.1mm}
  \begin{subfigure}[b]{0.11\textwidth}
      \centering
      \includegraphics[width=1\textwidth]{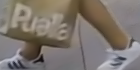}
      \caption{LoFormer}
  \end{subfigure}
  \hfill
  \vspace{0.1mm}
  \begin{subfigure}[b]{0.11\textwidth}
      \centering
      \includegraphics[width=1\textwidth]{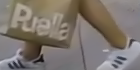}
      \caption{\bf Ours}
  \end{subfigure}
  \hfill
  \vspace{0.1mm}

  \begin{subfigure}[b]{0.11\textwidth}
    \centering
    \includegraphics[width=1\textwidth]{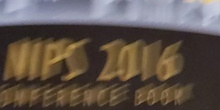}
    \caption{Blurred}
  \end{subfigure}
  \hfill
  \vspace{0.1mm}
  \begin{subfigure}[b]{0.11\textwidth}
      \centering
      \includegraphics[width=1\textwidth]{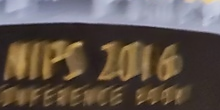}
      \caption{MPRNet}
  \end{subfigure}
  \hfill
  \vspace{0.1mm}
  \begin{subfigure}[b]{0.11\textwidth}
      \centering
      \includegraphics[width=1\textwidth]{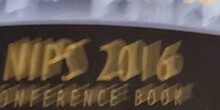}
      \caption{NAFNet}
  \end{subfigure}
  \hfill
  \vspace{0.1mm}
  \begin{subfigure}[b]{0.11\textwidth}
      \centering
      \includegraphics[width=1\textwidth]{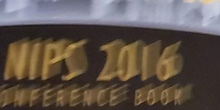}
      \caption{Restormer}
  \end{subfigure}
  \hfill
  \vspace{0.1mm}
  \begin{subfigure}[b]{0.11\textwidth}
      \centering
      \includegraphics[width=1\textwidth]{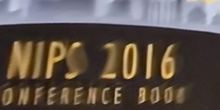}
      \caption{UFPNet}
  \end{subfigure}
  \hfill
  \vspace{0.1mm}
  \begin{subfigure}[b]{0.11\textwidth}
      \centering
      \includegraphics[width=1\textwidth]{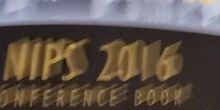}
      \caption{FFTformer}
  \end{subfigure}
  \hfill
  \vspace{0.1mm}
  \begin{subfigure}[b]{0.11\textwidth}
      \centering
      \includegraphics[width=1\textwidth]{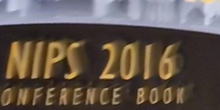}
      \caption{LoFormer}
  \end{subfigure}
  \hfill
  \vspace{0.1mm}
  \begin{subfigure}[b]{0.11\textwidth}
      \centering
      \includegraphics[width=1\textwidth]{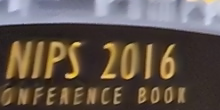}
      \caption{\bf Ours}
  \end{subfigure}
  \hfill
  \vspace{0.1mm}

  \setcounter{subfigure}{0}
  \begin{subfigure}[b]{0.11\textwidth}
    \centering
    \includegraphics[width=1\textwidth]{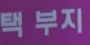}
    \caption{Reference}
  \end{subfigure}
  \hfill
  \vspace{0.1mm}
  \begin{subfigure}[b]{0.11\textwidth}
      \centering
      \includegraphics[width=1\textwidth]{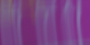}
      \caption{Blurred}
  \end{subfigure}
  \hfill
  \vspace{0.1mm}
  \begin{subfigure}[b]{0.11\textwidth}
      \centering
      \includegraphics[width=1\textwidth]{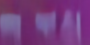}
      \caption{MPRNet}
  \end{subfigure}
  \hfill
  \vspace{0.1mm}
  \begin{subfigure}[b]{0.11\textwidth}
      \centering
      \includegraphics[width=1\textwidth]{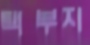}
      \caption{DeepRFT}
  \end{subfigure}
  \hfill
  \vspace{0.1mm}
  \begin{subfigure}[b]{0.11\textwidth}
      \centering
      \includegraphics[width=1\textwidth]{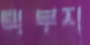}
      \caption{UFPNet}
  \end{subfigure}
  \hfill
  \vspace{0.1mm}
  \begin{subfigure}[b]{0.11\textwidth}
      \centering
      \includegraphics[width=1\textwidth]{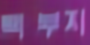}
      \caption{FFTformer}
  \end{subfigure}
  \hfill
  \vspace{0.1mm}
  \begin{subfigure}[b]{0.11\textwidth}
      \centering
      \includegraphics[width=1\textwidth]{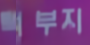}
      \caption{MISC}
  \end{subfigure}
  \hfill
  \vspace{0.1mm}
  \begin{subfigure}[b]{0.11\textwidth}
      \centering
      \includegraphics[width=1\textwidth]{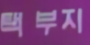}
      \caption{\bf Ours}
  \end{subfigure}
  \hfill
  \vspace{0.1mm}
    \caption{From the top to bottom rows: qualitative comparisons on the GoPro~\cite{nah2017deep}, HIDE~\cite{shen2019human}, RWBI~\cite{zhang2020deblurring} and RealBlur-J~\cite{rim2020real} testsets. For the first three row, the deblurring networks are trained only on the GoPro training set and directly applied to GoPro, HIDE and RWBI test sets. For the bottom row, we train and test models on the RealBlur-J dataset. Our XYScanNet produces sharper and visually pleasant results. Refer to the released codes for additional visual results.}\label{fig:sota_gopro}
\end{figure*}

 Unlike the asymmetric feature fusion (AFF)~\cite{cho2021rethinking} in~\cref{figure:ms} (a), which relies on traditional convolution designs, our proposed dual gating feature fusion (DGFF) introduces two feature-control gates. Each gate consists of a KL divergence term between the current-level and other-level features, followed by a sigmoid activation function. We incorporate KL divergence to enhance model interpretability by measuring the distributional distance between feature levels. The computed divergence is then passed through an activation function, allowing the model to selectively propagate visual information across multiple levels, as shown in~\cref{figure:ms}b. Given an input feature $\mathbf{X}_{cur}$ from the current level and the other two feature maps $\mathbf{X}_{oth1}$ and $\mathbf{X}_{oth2}$ from different levels, the output $\mathbf{\hat{X}}$ is defined as,
\begin{equation} 
  \begin{aligned} \label{eq:DGFF}
  \mathbf{\hat{X}} = &W_{p}\bigl\{\{\mathrm{\sigma}[\mathrm{KL}(\mathbf{X}^{\prime}_{oth1}, \mathbf{X}^{\prime}_{cur})]\odot\mathbf{X}^{\prime}_{oth1}+\\
  & \mathrm{\sigma}[\mathrm{KL}(\mathbf{X}^{\prime}_{oth2}, \mathbf{X}^{\prime}_{cur})]\odot\mathbf{X}^{\prime}_{oth2}\}||\mathbf{X}^{\prime}_{cur}\bigr\},\\
  \end{aligned}
\end{equation}
where $\mathbf{X}^{\prime}_{cur}$, $\mathbf{X}^{\prime}_{oth1}$, $\mathbf{X}^{\prime}_{oth2}$ denote the projections obtained from resizing and applying pixel-wise convolutions to their respective input features; $\mathrm{KL}(\mathit{y_{in}, y_{ref}})$ represent the KL-Divergence computed by $\mathit{y_{ref}\cdot(\mathrm{log}y_{ref}-\mathrm{log}y_{in})}$; and $||$ denotes feature concatenation.


\section{Experiments and Analysis}
\label{sec:experiment}


\subsection{Implementation Details}\label{sec:implementation_details}
 We train XYScanNet on the GoPro~\cite{nah2017deep} dataset with evaluation on GoPro, RWBI~\cite{zhang2020deblurring}, and HIDE~\cite{shen2019human} test sets. 
 For training details, please refer to our released codes,~{\url{https://github.com/HanzhouLiu/XYScanNet}}. Following prior work~\cite{kupyn2019deblurgan, tsai2022stripformer, liu2024deblurdinat}, the loss function is given as, 
    \begin{equation}
    L = L_{\text{char}} + \lambda_{1} L_{\text{edge}} + \lambda_{2} L_{\text{p}},
    \end{equation}
 where $L_{\text{char}}$ denotes the Charbonnier loss, $L_{\text{edge}}$ is the edge loss, $L_{\text{p}}$ represents the feature distance extracted by VGG, $\lambda_1 = 0.05$, and $\lambda_2 = 0.0005$. Pixel-pixel losses, $L_{\text{char}}$ and $L_{\text{edge}}$, align with distortion criteria, whereas $L_{\text{p}}$ is more aligned with human perception. According to previous studies~\cite{blau2018perception, blau2019rethinking, zhang2022perception, zhu2024perceptual}, weighted losses in different proportions do not aim to improve both losses simultaneously. {\bf While additional studies on different training strategies are planned, they are beyond the scope of this paper.} For fair comparisons, we evaluate deblurred results by both perceptual and distortion metrics. Training on RealBlur-J and RealBlur-R~\cite{rim2020real} follows the previous setup~\cite{fang2023self}.
\subsection{Comparisons with State-of-the-art Methods}
\begin{table}[tb]
  \setlength{\tabcolsep}{2.1pt}
  \centering
  \begin{tabular}{l cc cc | cc}
    \toprule
    \multirow{2}{*}{\bf Model} & \multicolumn{4}{c}{{\bf Perceptual} $\downarrow$} & \multicolumn{2}{c}{{\bf Distortion} $\uparrow$}\\
    \cmidrule(lr){2-5} \cmidrule(lr){6-7}
    & KID & FID & LPIPS & NIQE & PSNR & SSIM\\
    \midrule
    Ground Truth & 0.0 & 0.0 & 0.0 & 3.21 & $\infty$ & 1.000\\
    \midrule
    {\bf ACM MM} \textcolor{red}{24}\\
    ALGNet$^\dag$\footref{foot2}~\cite{gao2024aggregating} & 0.122 & 0.192 & 0.089 & 4.11 & 33.49 & 0.964 \\
    LoFormer~\cite{mao2024loformer} & 0.089 & 0.152 & 0.072 & \bf4.05 & 34.09 & \bf0.969\\
    \midrule
    {\bf CVPR} \textcolor{red}{21-24}\\
    MPRNet~\cite{zamir2021multi} & 0.114 & 0.204 & 0.089 & 4.09 & 32.66 & 0.959\\
    DSR-SA$^\ast$\footref{foot1}~\cite{whang2022deblurring} & - & - & 0.078 & \ul{4.07} & 33.23 & 0.963\\
    Restormer~\cite{zamir2022restormer} & 0.121 & 0.199 & 0.084 & 4.11 & 32.92 & 0.961\\ 
    UFPNet~\cite{fang2023self} & 0.089 & \ul{0.148} & 0.076 & \ul{4.07} & 34.06 & \ul{0.968}\\ 
    FFTformer~\cite{kong2023efficient} & 0.097 & 0.152 & \ul{0.071} & \bf4.05 & \bf34.21 & \bf0.969\\
    MISC$^\ast$\footref{foot1}~\cite{liu2024motion} & - & - & - & - & \ul{34.10} & \bf0.969\\
    \midrule
    {\bf ECCV} $\textcolor{red}{22}$\\
    NAFNet~\cite{chen2022simple} & \ul{0.088} & 0.157 & 0.078 & \ul{4.07} & 33.71 & 0.967\\ 
    \midrule
    {\bf Ours}\\
    \rowcolor{lightgray}XYScanNet & \bf 0.073 & \bf 0.138 & \bf 0.067 & \bf 4.05 & 33.91 & \ul{0.968}\\
    \bottomrule
  \end{tabular}
  \caption{
Deblurring results on GoPro~\cite{nah2017deep}. Numbers in $\textcolor{red}{\textrm{red}}$ indicate the publication year. XYScanNet outperforms recent deblurring models across all perceptual metrics, albeit with slightly worse distortion metrics. {\bf This performance can be attributed to the loss function discussed in~\cref{sec:implementation_details}.} KID and FID are implemented by~\cite{parmar2022aliased} and normalized to a range of $[0, 1]$.}\label{tab:gopro}
\end{table}

\subsubsection{Qualitative Analysis}

\begin{table*}[tb]
  \setlength{\tabcolsep}{2.5pt}
  \centering
  \begin{tabular}{l c c c c c c c >{\columncolor{lightgray}}c}
    \toprule
    \bf Method & Blurred & MPRNet~\cite{zamir2021multi} 
    &Restormer~\cite{zamir2022restormer}
    &NAFNet~\cite{chen2022simple} &UFPNet~\cite{fang2023self} &FFTformer~\cite{kong2023efficient} &LoFormer~\cite{mao2024loformer} & XYScanNet\\
    \midrule
    \bf NIQE$\downarrow$ & 3.546 & 3.591 &3.594 &3.441 &3.690 &3.384 &3.536 &\bf3.383\\
    \bottomrule
  \end{tabular}
  \caption{
Comparisons of GoPro-trained networks evaluated on the real-world RWBI~\cite{zhang2020deblurring} test set which contains no reference image.}\label{tab:rwbi}
\end{table*}

\cref{fig:sota_gopro} highlights the limitations of existing CNN and Transformer methods~\cite{zamir2021multi,chen2022simple,fang2023self,zamir2022restormer,kong2023efficient,mao2024loformer} in processing severely blurred images. These approaches fall short in restoring sharp edges, clear characters, and distinct patterns from such inputs. By contrast, our Mamba-based method effectively recovers these elements, producing images perceptually closer to the ground truth with noticeably enhanced visual quality. Furthermore, improved visual quality on HIDE~\cite{shen2019human} and RWBI~\cite{zhang2020deblurring} datasets demonstrate XYScanNet’s superior generalization capabilities. The bottom row of~\cref{fig:sota_gopro} proves that XYScanNet also shows high deblurring performance in real-world scenarios. More specifically, our proposed XYScanNet exhibits superior deblurring performance at the edges of local regions, particularly on the characters and geometric designs with high contrast against the background. This is attributed to an accurate understanding of pixel dependencies in blurred images.

\subsubsection{Quantitative Analysis}

\begin{table}[tb]
  \setlength{\tabcolsep}{2.1pt}
  \centering
  \begin{tabular}{l cc cc |cc}
    \toprule
    \multirow{2}{*}{\bf Model} & \multicolumn{4}{c}{\bf Perceptual $\downarrow$} & \multicolumn{2}{c}{\bf Distortion $\uparrow$}\\
    \cmidrule(lr){2-5} \cmidrule(lr){6-7}
    & KID & FID & LPIPS & NIQE & PSNR & SSIM\\
    \midrule
    Ground Truth & 0.0 & 0.0 & 0.0 & 2.72 & $\infty$ & 1.000\\
    \midrule
    MPRNet~\cite{zamir2021multi} & 0.055 & 0.168 & 0.114 & 3.46 & 30.96 & 0.939\\
    DSR-SA$^\ast$\footref{foot1}~\cite{whang2022deblurring} & - & - & \ul{0.092} & \bf 2.93 & 30.07 & 0.928\\
    NAFNet~\cite{chen2022simple} & 0.043 & 0.137 & 0.103 & 3.22 & 31.31 & 0.942\\ 
    Restormer~\cite{zamir2022restormer} & 0.052 & 0.150 & 0.108 & 3.42 & 31.22 & 0.942\\
    UFPNet~\cite{fang2023self} & 0.066 & 0.150 & 0.093 & \ul{3.13} & \ul{31.74} & \ul{0.947}\\ 
    FFTformer~\cite{kong2023efficient} & 0.051 & 0.139 & 0.096 & 3.29 & 31.62 & 0.945\\
    MISC$^\ast$\footref{foot1}~\cite{liu2024motion} & - & - & - & - & 31.66 & 0.946\\
    ALGNet$^\dag$\footref{foot2}~\cite{gao2024aggregating} & 0.083 & 0.166 & 0.100 & 3.20 & 31.64 & \ul{0.947} \\
    LoFormer~\cite{mao2024loformer} & \ul{0.040} & \ul{0.125} & 0.093 & 3.30 & \bf31.86 & \bf0.949\\
    \midrule
    \rowcolor{lightgray}XYScanNet & \bf 0.035 & \bf 0.118 & \bf 0.091 & 3.52 & \ul{31.74} & \ul{0.947}\\
    \bottomrule
  \end{tabular}
  \caption{
  Comparisons of GoPro-trained networks on HIDE~\cite{shen2019human}. Anomaly elevated NIQE$\downarrow$ is caused by a biased dataset focusing on human images according to previous studies~\cite{zvezdakova2019barriers}.
}\label{tab:hide}
\end{table}

In Table 1-5, the best results are {\bf bolded}, and the second-best results are \ul{underlined}. Refer to Footnotes~\footnotemark{} \footnotetext{$^\ast$ indicates results taken directly from the paper.\label{foot1}}\footnotemark{} \footnotetext{$^\dag$ denotes results tested on images provided by the authors.\label{foot2}}.

\medskip
\noindent
{\bf{GoPro results.}}
We train and evaluate XYScanNet on the GoPro dataset~\cite{nah2017deep}. The image quality scores of the deblurred results are summarized in~\cref{tab:gopro}, where XYScanNet consistently outperforms existing deblurring networks across all perceptual metrics while maintaining a competitive PSNR of $33.91$ dB and SSIM of $0.968$. 
Notably, XYScanNet achieves a KID score of 0.073, nearly a $25\%$ reduction compared to FFTformer~\cite{kong2023efficient} with the highest distortion metrics on the GoPro testset, demonstrating that our method achieves state-of-the-art deblurring performance in terms of perceptual metrics on GoPro.

\medskip
\noindent
{\bf{RWBI results.}}
We directly apply the GoPro-trained models on the RWBI dataset~\cite{zhang2020deblurring} which features real-world blur and contains no ground-truth image, metric scores of which are reported in~\cref{tab:rwbi}. MPRNet~\cite{zamir2021multi}, Restormer~\cite{zamir2022restormer}, and UFPNet~\cite{fang2023self} generate deblurred images with higher NIQE scores than the original blurred inputs, indicating reduced naturalness. In contrast, networks like NAFNet~\cite{chen2022simple}, LoFormer~\cite{mao2024loformer}, FFTformer~\cite{kong2023efficient}, and XYScanNet achieve lower NIQE scores, reflecting improved deblurring quality. XYScanNet achieves the highest NIQE reduction of 0.163, surpassing NAFNet’s 0.105, LoFormer’s 0.01, and performing competitively with FFTformer’s 0.162. It indicates the powerful generalization of XYScanNet to the images with real-world blurred patterns.

\medskip
\noindent
{\bf{HIDE results.}}
We further assess generalization capabilities of the GoPro-trained networks on the HIDE testset~\cite{shen2019human}, which primarily consists of human images, to assess their generalization capabilities to broader domains. As shown in~\cref{tab:hide}, XYScanNet achieves the second-highest PSNR of $31.74$ dB and SSIM of $0.947$, closely following the latest LoFormer~\cite{mao2024loformer}. While it has been reported that the fine facial textures in human centric images elevate NIQE values~\cite{zvezdakova2019barriers}, we report NIQE for thorough comparisons.~\cref{tab:hide} reports that XYScanNet surpasses recent networks across nearly all perceptual metrics except NIQE on HIDE, suggesting the effectiveness of our approach to process blurred images focused on human beings.

\begin{table}[tb]
  \setlength{\tabcolsep}{1.8pt}
  \centering
  \begin{tabular}{l cc | cc}
    \toprule
    \multirow{2}{*}{\bf Method} & \multicolumn{2}{c}{\bf RealBlur-J} & \multicolumn{2}{c}{\bf RealBlur-R}\\
    \cmidrule(lr){2-3} \cmidrule(lr){4-5}
    & Q-ALIGN$\uparrow$ & NIQE$\downarrow$ & Q-ALIGN$\uparrow$ & NIQE$\downarrow$\\
    \midrule
    MPRNet~\cite{zamir2021multi} & 0.550 & 4.11 & 0.238 & 5.59\\
    DeepRFT$^\dag$\footref{foot2}~\cite{mao2023intriguing} & 0.560 & 3.89 & 0.240 & 5.30\\
    UFPNet~\cite{fang2023self} & 0.572 & 3.94 & 0.245 & \ul{5.23}\\ 
    FFTformer~\cite{kong2023efficient} & 0.461 & 3.84 & 0.249 & 5.45\\
    LoFormer~\cite{mao2024loformer}  & \ul{0.592} & \bf3.71  & \ul{0.251} & 5.31\\
    MISC$^\dag$\footref{foot2}~\cite{liu2024motion} & 0.570 & \ul{3.80} & 0.243 & 5.45\\
    \midrule
    \rowcolor{lightgray}XYScanNet & \bf0.601 & \bf3.71 & \bf0.266 & \bf5.21\\
    \bottomrule
  \end{tabular}
  \caption{
Comparisons of recent models trained and tested on RealBlur datasets~\cite{rim2020real}. Q-ALIGN and NIQE are reported as no-reference metrics due to {\bf misaligned image pairs} in RealBlur.}\label{tab:realblur}
\end{table}

\begin{figure*}[tb]
  \centering
  \includegraphics[width=0.97\textwidth, trim = 0.4cm 4.3cm 2.8cm 0.4cm]{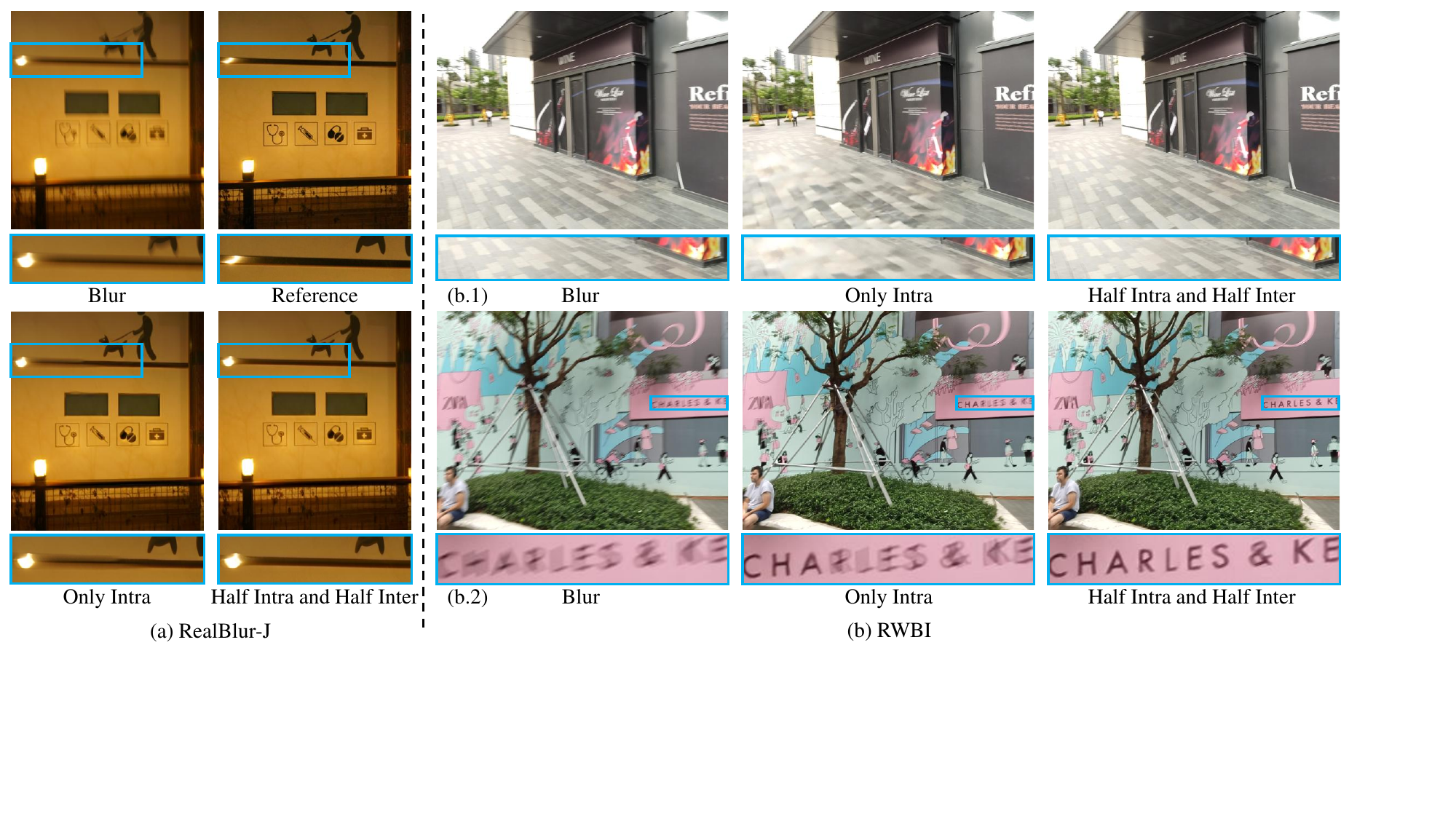}

  \caption{Replacing half of the Intra-Scanners with Inter-Scanners (refer to~\cref{figure:architecture}b) enhances texture capture across the image and improves neatly arranged character clarity on large billboards, confirming the necessity of Intra-VSSM for {\bf large-area blurred patterns}.}\label{figure:abl_inter}
\end{figure*}

\medskip
\noindent
{\bf{RealBlur results.}}
Additionally, we train XYScanNet on the RealBlur datasets~\cite{rim2020real}, which consist of {\bf misaligned} image pairs captured in real-world scenarios. For a fair comparison, we only employ image quality metrics without reference image needed, \ie, a large multi-modality model (LMM) -based metric, Q-ALIGN~\cite{wu2023q}, and a conventional metric NIQE.~\cref{tab:realblur} reports that {\bf XYScanNet surpasses all recent models} as rated by Q-ALIGN, indicating a closer alignment with human perception in the real-world scenarios. Moreover, the improved NIQE scores validate the enhanced naturalness of our XYScanNet’s deblurred results across multiple real-world scenes.


\subsection{Ablation Study}
\begin{figure*}[tb]
  \centering
  \includegraphics[width=0.94\textwidth, trim = 0.34cm 11.1cm 12.75cm 0.34cm]{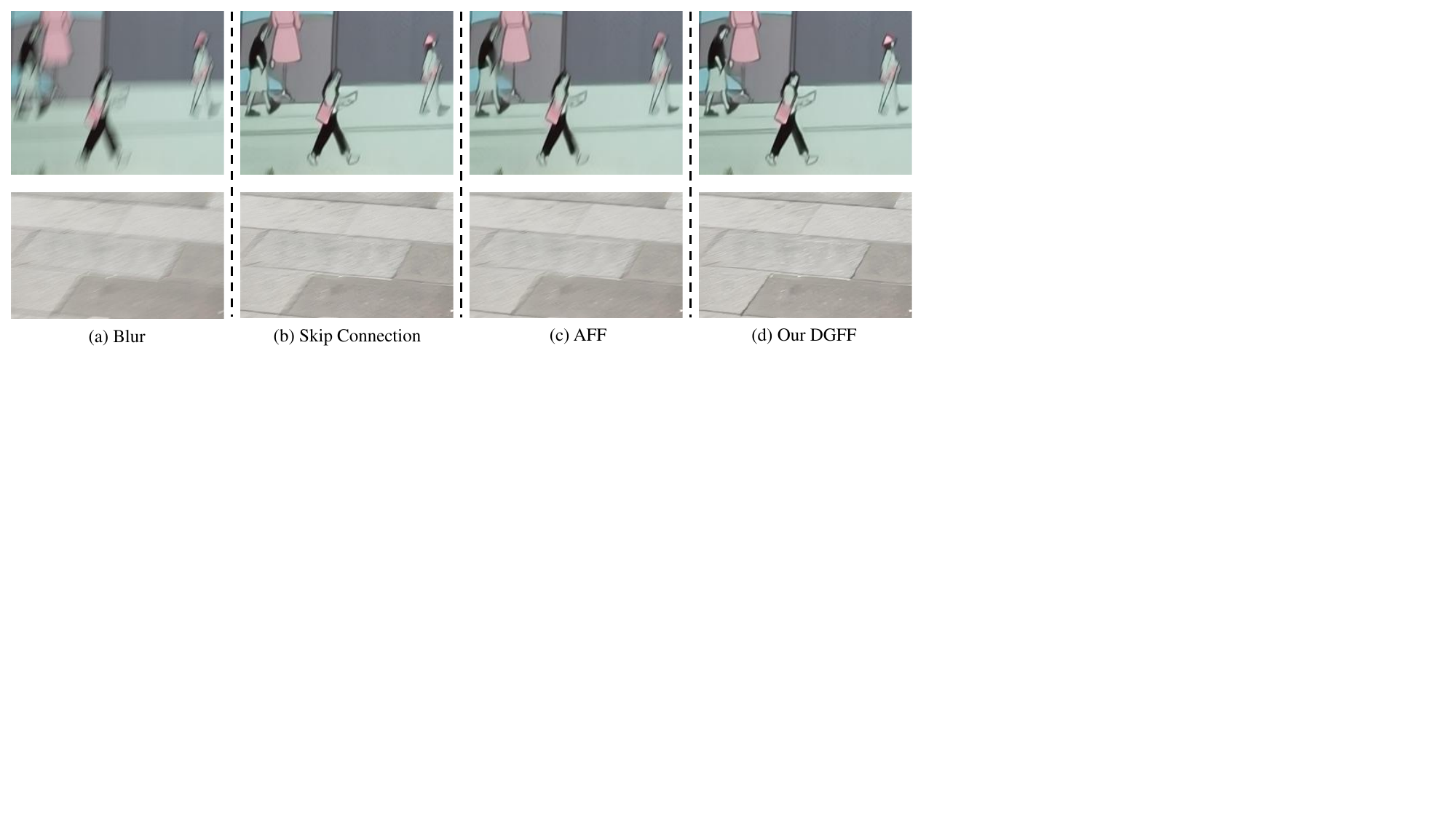}

  \caption{On real-world RWBI~\cite{zhang2020deblurring}, {\bf our DGFF} is able to estimate the blur patterns that are ignored by skip connections and AFF~\cite{cho2021rethinking}.}\label{fig:abl_dgff}
\end{figure*}

For the ablation studies, we train the networks on the GoPro dataset~\cite{nah2017deep} with image patches of size $128\times128$ for 3K epochs. We measure efficiency on a local RTX 3090 GPU.

\begin{table}[tb]
  \setlength{\tabcolsep}{0.12cm}
  \centering 
  \begin{tabular}{c  c  c  c | c  c  c  c}
    \toprule
    \multirow{2}{*}{\bf Intra} 
    & \multirow{2}{*}{\bf Inter} &\multicolumn{2}{c}{\bf PSNR$\uparrow$} & \multicolumn{2}{c}{\bf Time (s)} & \multicolumn{2}{c}{\bf Space (GB)}\\
    \cmidrule(lr){3-4} \cmidrule(lr){5-6} \cmidrule(lr){7-8}
    & & GoPro & HIDE & Train & Test & Train & Test\\
    \midrule
    & \checkmark & 31.91 & 30.01 & \bf65 & \bf0.153 & \bf10.36 & \bf3.82\\
    \checkmark &  & {\bf32.18} & {\bf30.13} & 103 & 0.167 & 12.82 & 4.67\\
    \rowcolor{lightgray} \bf \checkmark & \checkmark & \underline{32.11} & \underline{30.11} & \underline{82} & \underline{0.165} & \underline{11.60} & \underline{4.51}\\
  \bottomrule
  \end{tabular}
  \caption{Replacing half of the Intra-Scanners with Inter-Scanners (refer to~\cref{figure:architecture}b) {\bf reduces computation cost} and {\bf improves global blur estimation} (see~\cref{figure:abl_inter}), with a slight drop in metrics.}\label{tab:abl_slice}
\end{table}

\begin{table}[tb]
  \setlength{\tabcolsep}{0.083cm}
  \centering 
  \begin{tabular}{c  c  c  c  c  c | c  c}
    \toprule
    \multirow{2}{*}{\bf AFF}
    & \multirow{2}{*}{\bf DGFF} &\multicolumn{2}{c}{{\bf NIQE$\downarrow$}} &\multicolumn{2}{c}{\bf LPIPS$\downarrow$} & \multicolumn{1}{c}{\bf Flops} & \multicolumn{1}{c}{\bf Params}\\
    \cmidrule(lr){3-4} \cmidrule(lr){5-6}
    &  & HIDE & RWBI & HIDE & RB.J& (G) & (K)\\
    \midrule
    & & 3.46 &\underline{3.52} & 0.125 & 0.889 & - & -\\ 
    \checkmark &  & {\bf3.40} &3.53 & \underline{0.124} & \underline{0.878} & 4.32 & 152.35\\
    \rowcolor{lightgray} & \checkmark & \underline{3.41} &{\bf3.51} & {\bf0.122} & \bf0.848 & {\bf2.19} & {\bf71.43}\\       
  \bottomrule
  \end{tabular}
  \caption{Effects of cross-level feature fusion methods, AFF~\cite{cho2021rethinking} and {\bf our DGFF} modules. See~\cref{fig:abl_dgff} for visual comparisons.}\label{tab:abl_ff}
\end{table}

\medskip
\noindent
{\bf{Investigation on Intra- and Inter-VSSM.}}
We set the total number of VSSMs as $K$ in all models. The interleaved approach contains $\frac{K}{2}$ Intra-VSSMs and $\frac{K}{2}$ Inter-VSSMs.~\cref{tab:abl_slice} reports that the interleaved method improves PSNR on GoPro~\cite{nah2017deep} by $0.2$ dB over the inter-only method. Compared to the intra-only approach,~\cref{figure:abl_inter} demonstrates improved visual fidelity of the interleaved approach, with a $20.39\%$ reduction in training time and a $9.52\%$ decrease in memory usage as shown in~\cref{tab:abl_slice}.

\begin{table*}[tb]
  \setlength{\tabcolsep}{2pt}
  \centering
  
  \begin{tabular}{l | ccccc | ccccc | c | c | cc | cc}
    \toprule
    \multirow{2}{*}{\bf Method} & \multicolumn{5}{c}{\bf GoPro} & \multicolumn{5}{c}{\bf HIDE} & \multicolumn{1}{c}{\bf RWBI} & \multicolumn{1}{c}{\bf\#P.} & \multicolumn{2}{c}{\bf Training} & \multicolumn{2}{c}{\bf Inference}\\
    \cmidrule(lr){2-6} \cmidrule(lr){7-11} \cmidrule(lr){14-15} \cmidrule(lr){16-17}
    & PSNR & KID & FID & LPIPS & NIQE & PSNR & KID & FID & LPIPS & NIQE & NIQE & (M) & T (s) & S (GB) & T (s) & S (GB)\\
    \midrule
    Reference & $\infty$ & 0.0 & 0.0 & 0.0 & 3.21 & $\infty$ & 0.0 & 0.0 & 0.0 & 2.72 &-&- &- &- &- &-\\
    \midrule
    MambaIR & 31.94 &\bf0.121 &0.232 & 0.094 & 4.06 & \bf30.22 &0.074 &0.201 &\bf0.121 & 3.46 &3.63 &7.41 &189 &17.39 &0.196 &8.38\\
    \rowcolor{lightgray}Ours & \bf32.11 &0.129 &\bf0.210 &\bf0.091 & 4.06 & 30.11 &\bf0.064 &\bf0.188 & 0.122 &\bf3.41 &\bf3.51 &\bf7.30 &\bf82 &\bf11.60 &\bf0.165 &\bf4.51\\
   \bottomrule
  \end{tabular}
  \caption{Comparisons of the vision state space module from MambaIR~\cite{guo2024mambair} and ours with the same baseline framework, training strategy and similar network size. Our method surpasses the previous one~\cite{guo2024mambair} on mulitple metrics with significantly improved efficiency. {\bf Visual comparisons in~\cref{fig:abl_vssm} and ERF visualization in~\cref{fig:erf_abl} further validate our contribution.} T denotes time and S represents space.}\label{tab:mambair}
\end{table*}

\medskip
\noindent
{\bf{Effect of cross-level feature fusion.}}
\cref{fig:abl_dgff} demonstrates the effectiveness of our DGFF in recovering local details in real-world scenarios with domain adaptations.~\cref{tab:abl_ff} shows that DGFF reduces NIQE by $0.05$ on HIDE~\cite{shen2019human} and LPIPS$^{st}$ by $0.041$ on RealBlur-J~\cite{rim2020real} over conventional skip connections. Additionally, compared to the AFF method~\cite{cho2021rethinking}, our DGFF reduces FLOPS by $\mathbf{49.3\%}$ and Params by $\mathbf{53.11\%}$, demonstrating the improved efficiency.

\begin{figure}[tb]
  \captionsetup[subfigure]{labelformat=empty, justification=centering}
  \centering
  \begin{subfigure}[b]{0.11\textwidth}
    \centering
    \includegraphics[width=1\textwidth]{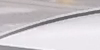}
  \end{subfigure}
  \hfill
  \vspace{0.5mm}
  \begin{subfigure}[b]{0.11\textwidth}
      \centering
      \includegraphics[width=1\textwidth]{figure/ablation/gopro/GOPR0410_11_00/cropped_gt.png}
  \end{subfigure}
  \hfill
  \vspace{0.5mm}
  \begin{subfigure}[b]{0.11\textwidth}
      \centering
      \includegraphics[width=1\textwidth]{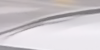}
  \end{subfigure}
  \hfill
  \vspace{0.5mm}
  \begin{subfigure}[b]{0.11\textwidth}
      \centering
      \includegraphics[width=1\textwidth]{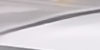}
  \end{subfigure}
  \hfill
  \vspace{0.5mm}

  \begin{subfigure}[b]{0.11\textwidth}
    \centering
    \includegraphics[width=1\textwidth]{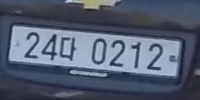}
  \end{subfigure}
  \hfill
  \vspace{0.5mm}
  \begin{subfigure}[b]{0.11\textwidth}
      \centering
      \includegraphics[width=1\textwidth]{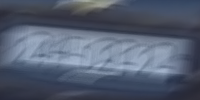}
  \end{subfigure}
  \hfill
  \vspace{0.5mm}
  \begin{subfigure}[b]{0.11\textwidth}
      \centering
      \includegraphics[width=1\textwidth]{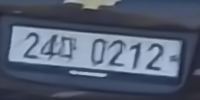}
  \end{subfigure}
  \hfill
  \vspace{0.5mm}
  \begin{subfigure}[b]{0.11\textwidth}
      \centering
      \includegraphics[width=1\textwidth]{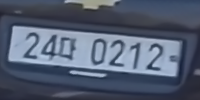}
  \end{subfigure}
  \hfill
  \vspace{0.5mm}

  \begin{subfigure}[b]{0.11\textwidth}
    \centering
    \includegraphics[width=1\textwidth]{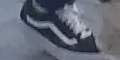}
    \caption{Reference}
  \end{subfigure}
  \hfill
  \vspace{0.5mm}
  \begin{subfigure}[b]{0.11\textwidth}
      \centering
      \includegraphics[width=1\textwidth]{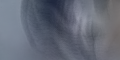}
      \caption{Blurred}
  \end{subfigure}
  \hfill
  \vspace{0.5mm}
  \begin{subfigure}[b]{0.11\textwidth}
      \centering
      \includegraphics[width=1\textwidth]{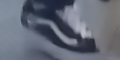}
      \caption{MambaIR}
  \end{subfigure}
  \hfill
  \vspace{0.5mm}
  \begin{subfigure}[b]{0.11\textwidth}
      \centering
      \includegraphics[width=1\textwidth]{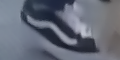}
      \caption{\bf Ours}
  \end{subfigure}
  \hfill
  \caption{Qualitative comparisons of the proposed VSSM versus the previous one in MambaIR~\cite{guo2024mambair}. 
  Our method generates objects with sharper and clearer boundaries, suggesting improvements in mitigating the limitations of existing Mamba work.}\label{fig:abl_vssm}
\end{figure}

\begin{figure}[tb]
  \captionsetup[subfigure]{labelformat=parens, justification=centering}
  \centering
  \begin{subfigure}[b]{0.22\textwidth}
      \centering
      \includegraphics[width=0.9\textwidth, trim = 3.5cm 2.8cm 3.2cm 2.9cm]{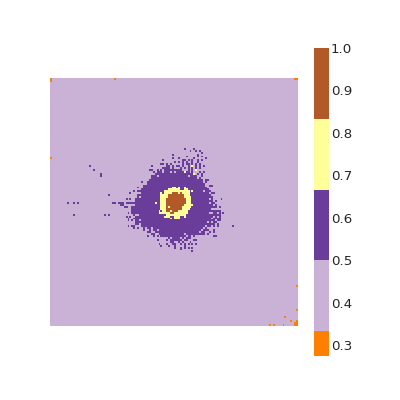}
      \caption{Ours.}
  \end{subfigure}
  \hfill
  \begin{subfigure}[b]{0.22\textwidth}
    \centering
    \includegraphics[width=0.9\textwidth, trim = 3.5cm 2.8cm 3.2cm 2.9cm]{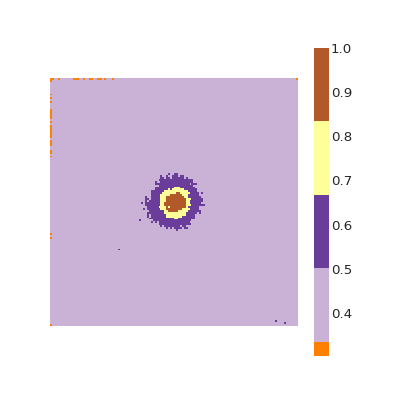}
    \caption{MambaIR.}
  \end{subfigure}
  \hfill
  \caption{Effective receptive fields (ERFs) of XYScanNet with (a) our VSSM and (b) the previous one proposed in MambaIR~\cite{guo2024mambair}. Ours has a sparser ERF with more intensive pixels.}\label{fig:erf_abl}
\end{figure}

\medskip
\noindent
{\bf{Analysis on vision state space modules.}}
The focus of this paper is to propose an efficient Vision State Space Module (VSSM) that outperforms the previous counterpart in MambaIR~\cite{guo2024mambair} on single-image deblurring tasks. To validate this contribution, we design ablation experiments conducted within the same baseline, ensuring that all components except the VSSM remain consistent. Visual results in~\cref{fig:abl_vssm} demonstrate that the previous VSSM~\cite{guo2024mambair} struggles to restore fine local details, even with cross-level feature fusion, due to the limitations of the flatten-and-scan approach, where {\bf spatial misalignment} (see~\cref{figure:scan}a) exists in each individual scanning route. Although brute-force quadrupling of scanning routes~\cite{guo2024mambair} implicitly address this issue, ~\cref{tab:mambair} shows that it is not a computation-friendly solution. In contrast, the proposed VSSM reaches a significant better balance between metric scores and computational costs, reducing the training time by $\mathbf{56.61\%}$ and inference memory usage by $\mathbf{46.18\%}$. The effectiveness of our VSSM is further verified by a larger receptive field as shown in~\cref{fig:erf_abl}.

\section{Conclusion}
In this work, we present a novel deep state space model, termed as XYScanNet, which effectively captures motion blur with varying magnitutes and angles. Specifically, we identify a long-standing issue, spatial misalignment, caused by the flatten-and-scan strategies in previous Mamba-based vision networks. To address this, we propose a novel slice-and-scan mechanism. The slicing process first divides image features into intra- and inter-slices. After that, selective scans are performed to capture pixel-wise dependencies within each intra-slice and slice-wise relationships across inter-slices. Building on this approach, we design a new Vision State Space Module (VSSM) demonstrating significantly improved efficiency compared to the previous VSSM. The proposed XYScanNet, with interleaved Intra- and Inter-VSSMs, is capable of recovering both fine local details and estimating large-area blur. Experimental results on multiple synthetic and real-world datasets verify that XYScanNet achieves state-of-the-art perceptual quality and competitive distortion metrics across diverse scenes.

\medskip
\noindent
{\bf Limitation.} 
First, XYScanNet may not achieve an optimal balance between local and global blur estimation. Second, further discussions are needed to explore the relationship between loss functions and performance. Lastly, extending XYScanNet to other restoration tasks presents a promising direction. These areas will be key focuses for future work.

\medskip
\noindent{\bf Acknowledgment.} We appreciate the supports of Texas A\&M High Performance Research Computing (HPRC) for providing access to the Grace and FASTER GPU clusters.

{
    \small
    \bibliographystyle{ieeenat_fullname}
    \bibliography{main}
}

\clearpage
\setcounter{page}{1}

\maketitlesupplementary

\section{Interpretability}
In this section, we outline the overall architecture of XYScanNet at first. After that, we explain the basic building blocks with ablation experiments. 

\subsection{Overall Pipeline}
As shown in~\cref{figure:architecture}, XYScanNet is of an asymmetric U-Net structure with cross-level (multi-scale) feature fusion. Given a blurred image $\mathbf{I_b}\in \mathbb{R}^{H\times W\times3}$, XYScanNet first uses an embedding layer with 2D convolutions to obtain shallow feature maps $\mathbf{F_s}\in \mathbb{R}^{H\times W\times C}$, where $H\times W$ represents the height by width, and $C$ denotes the number of hidden channels. Afterward, the low-level features $\mathbf{F_s}$ are processed through a three-level asymmetric encoder-decoder, where deep features $\mathbf{F_d}\in \mathbb{R}^{H\times W\times C}$ are learned. In the encoder path, only feed-forward networks exist. Finally, a convolution layer is applied to the refined features to generate residual image $\mathbf{I_R}\in \mathbb{R}^{H\times W\times3}$ to which degraded image is added to obtain the restored image: $\mathbf{I_c} = \mathbf{I_b} + \mathbf{I_R}$. We set the number of hidden channels $C=144$, and encoder or decoder blocks at each level $[N_1, N_2, N3]=[3, 3, 6]$.

\subsection{Individual Component}
In this subsection, we present an additional analysis of several basic components of XYScanNet.
As mentioned in the main manuscript,~\cref{tab:abl_slice} shows that the introduction of Intra-Scanners significantly improves training efficiency, reducing training time by $20.39\%$ and GPU usage by $1.22$ GB on GoPro~\cite{nah2017deep} with only 0.02 dB PSNR reduction on HIDE~\cite{shen2019human}. This finding reveals that this hybrid approach makes XYScanNet efficiently trained, and the resulting model is able to handle unknown blurry images despite minor metric score drop.

\medskip
\noindent
{\bf VSSM} proposed in this paper generates a sparser while darker effective receptive field (ERF) compared to the VSSM proposed in MambaIR~\cite{guo2024mambair}, as shown in~\cref{fig:erf_abl}. As shown in~\cref{tab:mambair}, our slice-and-scan-based VSSM.

\begin{figure}[tb]
  \centering
  \includegraphics[width=0.464\textwidth, trim = 0.5cm 13.6cm 20.8cm 0.4cm]{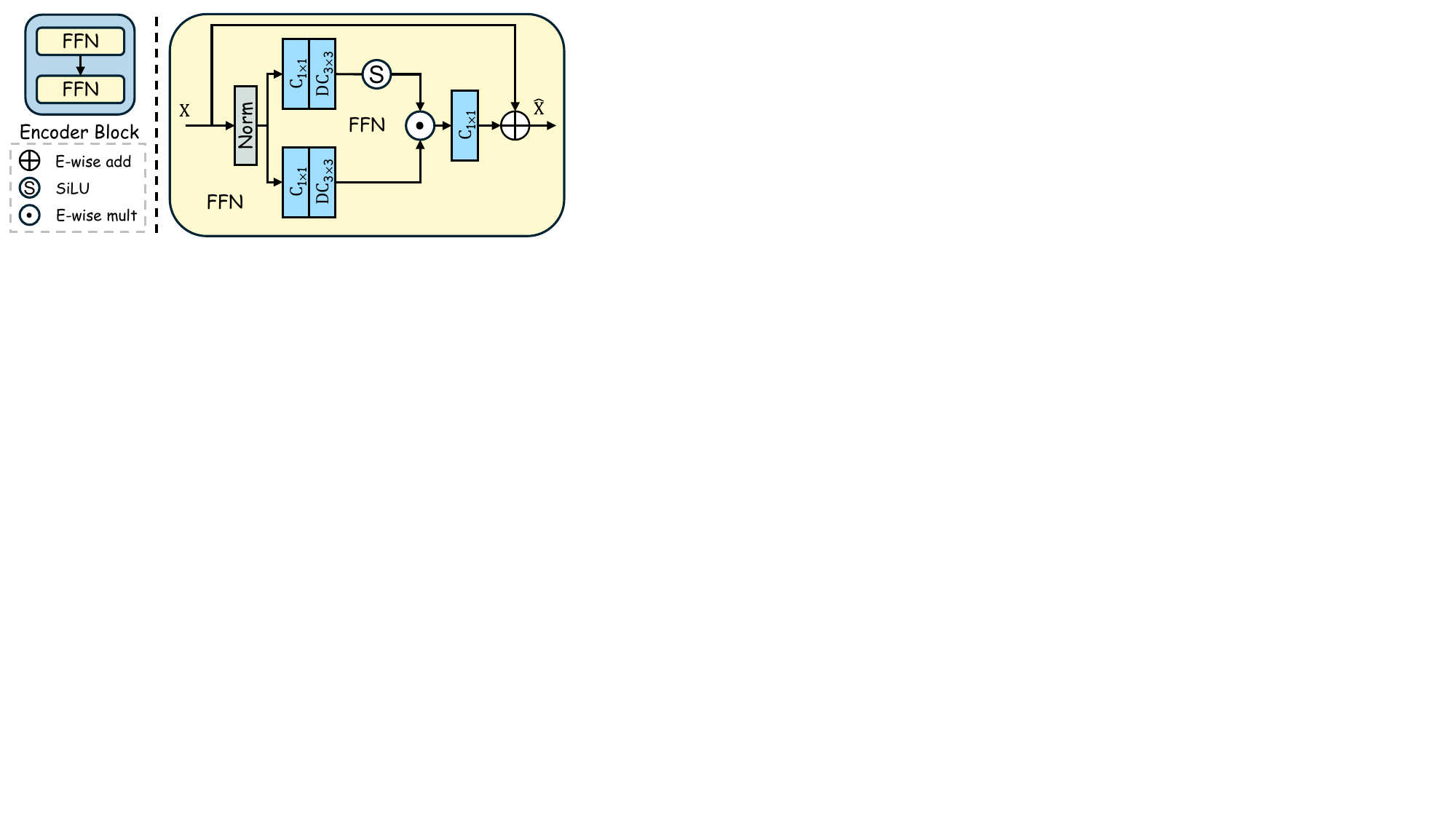}

  \caption{Supplementary to~\cref{figure:architecture}. Gated depth-wise feed-forward network~\cite{zamir2022restormer} in XYScanNet. We employ SiLU instead of GELU to keep consistent with activations in SSMs.}\label{figure:ffn}
\end{figure}

\medskip
\noindent
{\bf Feed-forward Networks} (FFNs) have been effectively studied in Transformer studies~\cite{zamir2022restormer,kong2023efficient,liu2024deblurdinat}. As shown in~\cref{figure:ffn}, we employ the Gated Depth-wise Feed-forward Network (GDFN)~\cite{zamir2022restormer} with SiLU instead of GELU to keep consistent with activations in mainstream State Space Models (SSMs). As stated in~\cite{zamir2022restormer}, GDFN introduces non-linearity to the model and controls the information propagation through the network, resulting in feature learning enriched with contextual knowledge.

\begin{figure}[tb]
  \centering
  \includegraphics[width=0.45\textwidth, trim = 0.5cm 7.6cm 19.7cm 0.4cm]{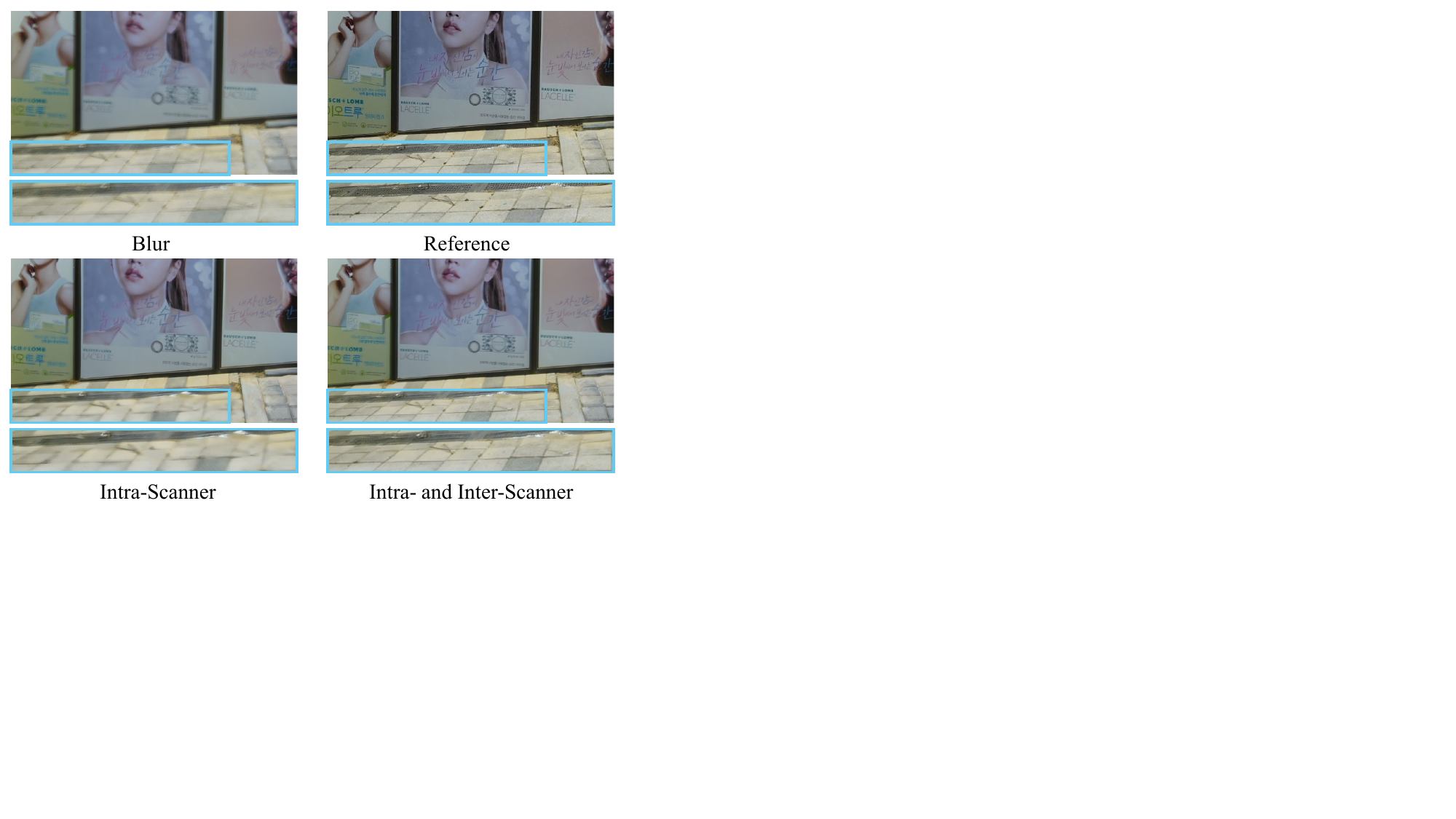}

  \caption{The hybrid method captures global and uniform blurred patterns (\eg, edges spanning across images) missed by the intra-only approach, despite minor metric score drop (see~\cref{tab:abl_slice}). The above images are cropped along the height by half to fit into the current page.}\label{figure:sup_inter}
\end{figure}

\section{Experiment Details}

\subsection{Experimental Settings}
We train the full XYScanNet on 8 A100 GPUs, each with 40 GB memory. To computer the metric scores, we use a local RTX 3090 GPU. Specifically, we calculate PSNR and SSIM with Matlab functions, while the other metrics by Python.  

\begin{figure*}[tb]
  \captionsetup[subfigure]{labelformat=parens, justification=centering}
  \centering
  \begin{subfigure}[b]{0.4\textwidth}
    \centering
    \includegraphics[width=1\textwidth]{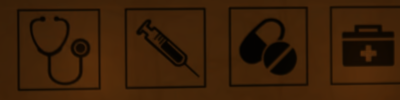}
    \caption{Reference}
  \end{subfigure}
  \hfill
  \begin{subfigure}[b]{0.4\textwidth}
      \centering
      \includegraphics[width=1\textwidth]{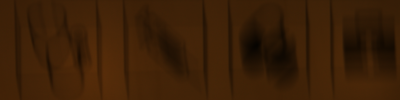}
      \caption{Blurred}
  \end{subfigure}
  \hfill
  \hfill
  \begin{subfigure}[b]{0.4\textwidth}
      \centering
      \includegraphics[width=1\textwidth]{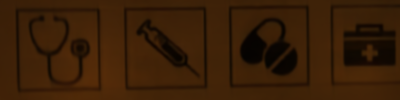}
      \caption{DeepRFT~\cite{mao2023intriguing}}
  \end{subfigure}
  \hfill
  \begin{subfigure}[b]{0.4\textwidth}
      \centering
      \includegraphics[width=1\textwidth]{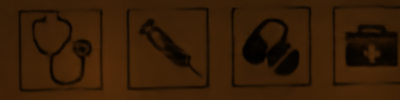}
      \caption{UFPNet~\cite{fang2023self}}
  \end{subfigure}
  \hfill
  \begin{subfigure}[b]{0.4\textwidth}
      \centering
      \includegraphics[width=1\textwidth]{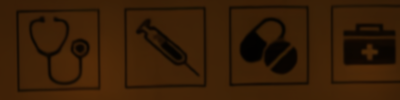}
      \caption{FFTformer~\cite{kong2023efficient}}
  \end{subfigure}
  \hfill
  \begin{subfigure}[b]{0.4\textwidth}
      \centering
      \includegraphics[width=1\textwidth]{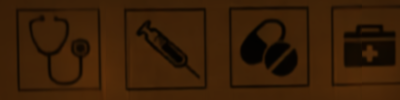}
      \caption{MISC~\cite{liu2024motion}}
  \end{subfigure}
  \hfill
  \begin{subfigure}[b]{0.4\textwidth}
      \centering
      \includegraphics[width=1\textwidth]{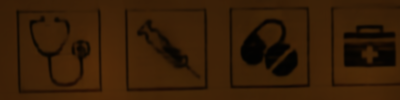}
      \caption{LoFormer~\cite{mao2024loformer}}
  \end{subfigure}
  \hfill
  \begin{subfigure}[b]{0.4\textwidth}
      \centering
      \includegraphics[width=1\textwidth]{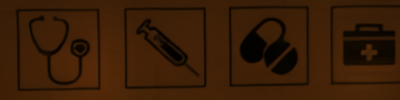}
      \caption{\bf Ours}
  \end{subfigure}
    \caption{Qualitative results of deblurring models trained and tested on the RealBlur-R dataset~\cite{rim2020real}.}\label{fig:sota_sup_rbr_1}
\end{figure*}

\medskip
\noindent
{\bf Ablation settings} are slightly different from the above. We only train the small version of XYScanNet with a hidden channel number $C=48$ on two A100 GPUs for an efficiency purpose. Each network has been trained and tested four times to minimize the effects of randomness. We conduct efficiency comparisons on a local RTX 3090 GPU for Input/Ouput stability. The training time is measured as seconds per epoch, while test time is computed as seconds per image. If not specified, the default dataset for training and inference is the GoPro dataset~\cite{nah2017deep}.

\subsection{Evaluation Metrics}
\label{sec:eval_metrics}
Different from mainstream deblurring work only computing PSNR and SSIM metrics, we use a wide range of metrics to quantify deblurring performance following~\cite{whang2022deblurring,liu2024deblurdinat}.

\medskip
\noindent
{\bf FID} (Fréchet Inception Distance)~\cite{heusel2017gans} quantifies the image quality by computing the Fréchet distance between feature distributions of the reference and output images, obtained from an Inception network.

\medskip
\noindent
{\bf KID} (Kernel Inception Distance)~\cite{binkowski2018demystifying} is used to measure the similarity between the ground truth and the generated images by comparing feature embeddings extracted from a pre-trained Inception network.

\medskip
\noindent
{\bf LPIPS}~\cite{zhang2019deep} evaluates perceptual similarity between two image patches by deep networks, which is more aligned with human perception of visual differences.

\medskip
\noindent
{\bf ST-LPIPS}~\cite{ghildyal2022stlpips} is suitable for RealBlur datasets~\cite{rim2020real} which contain misaligned image pairs, as explained in~\cite{liu2024deblurdinat}.

\medskip
\noindent
{\bf NIQE}~\cite{mittal2012making} is a non-deep learning image quality metric that measures the naturalness of an image based on its deviation from statistics, without requiring reference images. Experimental results in~\cite{zvezdakova2019barriers} reveal that the fine textures such as facial details elevate NIQE values~\cite{zvezdakova2019barriers}, making NIQE a questionable metric for the human-centric HIDE dataset~\cite{shen2019human}.

\medskip
\noindent
{\bf Q-ALIGN}~\cite{wu2023q} is a large multi-modality model (LMM) -based metric, which is considered as a emerging perceptual metric to evalute image quality.

\section{Performance Analysis}

\begin{figure*}[tb]
  \captionsetup[subfigure]{labelformat=parens, justification=centering}
  \centering
  \begin{subfigure}[b]{0.49\textwidth}
    \centering
    \includegraphics[width=1\textwidth]{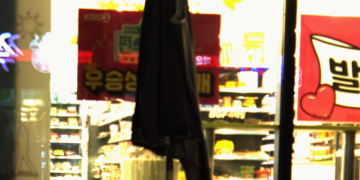}
    \caption{Reference}
  \end{subfigure}
  \hfill
  \begin{subfigure}[b]{0.49\textwidth}
      \centering
      \includegraphics[width=1\textwidth]{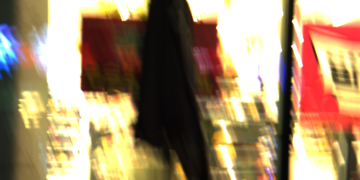}
      \caption{Blurred}
  \end{subfigure}
  \hfill
  \begin{subfigure}[b]{0.49\textwidth}
      \centering
      \includegraphics[width=1\textwidth]{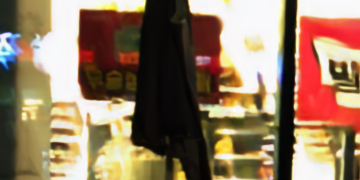}
      \caption{MPRNet~\cite{zamir2021multi}}
  \end{subfigure}
  \hfill
  \begin{subfigure}[b]{0.49\textwidth}
      \centering
      \includegraphics[width=1\textwidth]{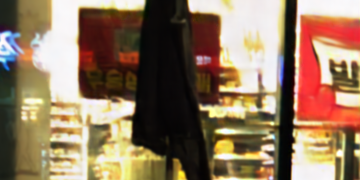}
      \caption{DeepRFT~\cite{mao2023intriguing}}
  \end{subfigure}
  \hfill
  \begin{subfigure}[b]{0.49\textwidth}
      \centering
      \includegraphics[width=1\textwidth]{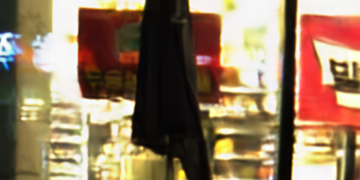}
      \caption{UFPNet~\cite{fang2023self}}
  \end{subfigure}
  \hfill
  \begin{subfigure}[b]{0.49\textwidth}
      \centering
      \includegraphics[width=1\textwidth]{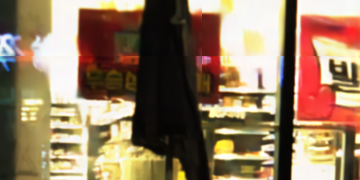}
      \caption{MISC~\cite{liu2024motion}}
  \end{subfigure}
  \hfill
  \begin{subfigure}[b]{0.49\textwidth}
      \centering
      \includegraphics[width=1\textwidth]{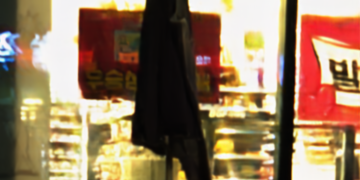}
      \caption{LoFormer~\cite{mao2024loformer}}
  \end{subfigure}
  \hfill
  \begin{subfigure}[b]{0.49\textwidth}
      \centering
      \includegraphics[width=1\textwidth]{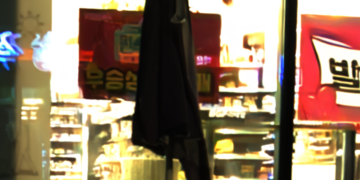}
      \caption{\bf Ours}
  \end{subfigure}
    \caption{Qualitative results of deblurring models trained and tested on RealBlur-R~\cite{rim2020real}.}\label{fig:sota_sup_rbr_2}
\end{figure*}

\begin{figure*}[tb]
  \captionsetup[subfigure]{labelformat=parens, justification=centering}
  \centering
  \begin{subfigure}[b]{0.49\textwidth}
    \centering
    \includegraphics[width=1\textwidth]{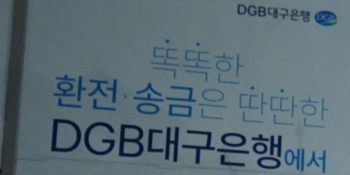}
    \caption{Reference}
  \end{subfigure}
  \hfill
  \begin{subfigure}[b]{0.49\textwidth}
      \centering
      \includegraphics[width=1\textwidth]{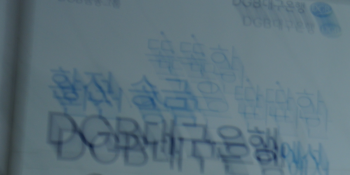}
      \caption{Blurred}
  \end{subfigure}
  \hfill
  \begin{subfigure}[b]{0.49\textwidth}
      \centering
      \includegraphics[width=1\textwidth]{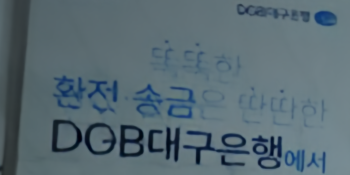}
      \caption{MPRNet~\cite{zamir2021multi}}
  \end{subfigure}
  \hfill
  \begin{subfigure}[b]{0.49\textwidth}
      \centering
      \includegraphics[width=1\textwidth]{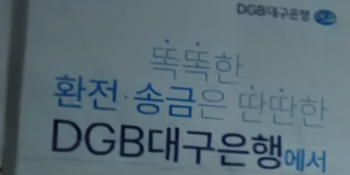}
      \caption{DeepRFT~\cite{mao2023intriguing}}
  \end{subfigure}
  \hfill
  \begin{subfigure}[b]{0.49\textwidth}
      \centering
      \includegraphics[width=1\textwidth]{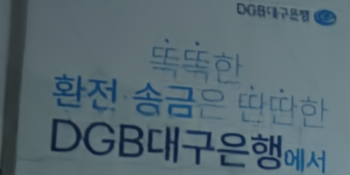}
      \caption{UFPNet~\cite{fang2023self}}
  \end{subfigure}
  \hfill
  \begin{subfigure}[b]{0.49\textwidth}
      \centering
      \includegraphics[width=1\textwidth]{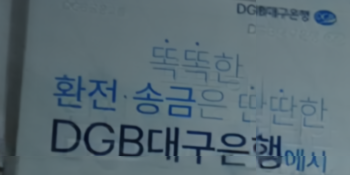}
      \caption{MISC~\cite{liu2024motion}}
  \end{subfigure}
  \hfill
  \begin{subfigure}[b]{0.49\textwidth}
      \centering
      \includegraphics[width=1\textwidth]{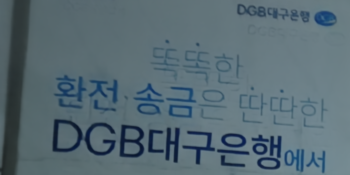}
      \caption{LoFormer~\cite{mao2024loformer}}
  \end{subfigure}
  \hfill
  \begin{subfigure}[b]{0.49\textwidth}
      \centering
      \includegraphics[width=1\textwidth]{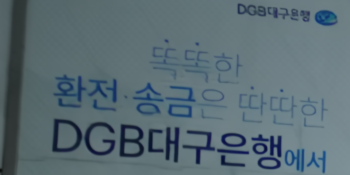}
      \caption{\bf Ours}
  \end{subfigure}
    \caption{Qualitative results of deblurring models trained and tested on RealBlur-J~\cite{rim2020real}.}\label{fig:sota_sup_rbj_1}
\end{figure*}

\begin{figure*}[tb]
  \captionsetup[subfigure]{labelformat=parens, justification=centering}
  \centering
  \begin{subfigure}[b]{0.49\textwidth}
    \centering
    \includegraphics[width=1\textwidth]{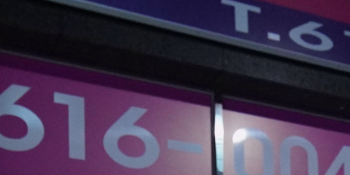}
    \caption{Reference}
  \end{subfigure}
  \hfill
  \begin{subfigure}[b]{0.49\textwidth}
      \centering
      \includegraphics[width=1\textwidth]{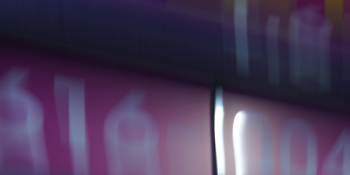}
      \caption{Blurred}
  \end{subfigure}
  \hfill
  \begin{subfigure}[b]{0.49\textwidth}
      \centering
      \includegraphics[width=1\textwidth]{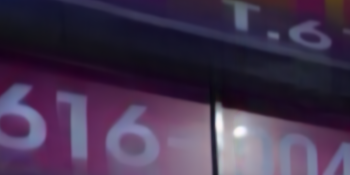}
      \caption{DeepRFT~\cite{mao2023intriguing}}
  \end{subfigure}
  \hfill
  \begin{subfigure}[b]{0.49\textwidth}
      \centering
      \includegraphics[width=1\textwidth]{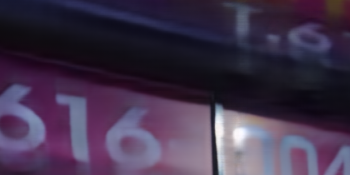}
      \caption{UFPNet~\cite{fang2023self}}
  \end{subfigure}
  \hfill
  \begin{subfigure}[b]{0.49\textwidth}
      \centering
      \includegraphics[width=1\textwidth]{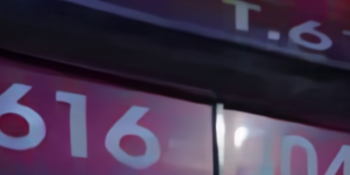}
      \caption{FFTformer~\cite{kong2023efficient}}
  \end{subfigure}
  \hfill
  \begin{subfigure}[b]{0.49\textwidth}
      \centering
      \includegraphics[width=1\textwidth]{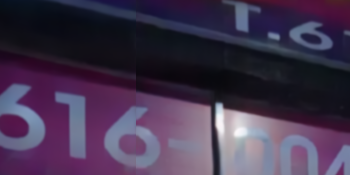}
      \caption{MISC~\cite{liu2024motion}}
  \end{subfigure}
  \hfill
  \begin{subfigure}[b]{0.49\textwidth}
      \centering
      \includegraphics[width=1\textwidth]{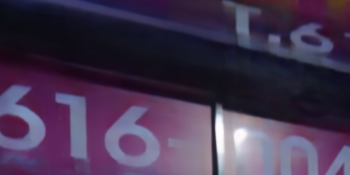}
      \caption{LoFormer~\cite{mao2024loformer}}
  \end{subfigure}
  \hfill
  \begin{subfigure}[b]{0.49\textwidth}
      \centering
      \includegraphics[width=1\textwidth]{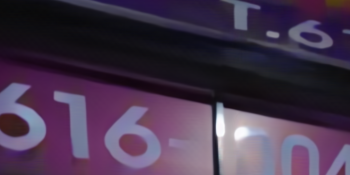}
      \caption{\bf Ours}
  \end{subfigure}
    \caption{Qualitative results of deblurring models trained and tested on RealBlur-J~\cite{rim2020real}.}\label{fig:sota_sup_rbj_2}
\end{figure*}

\begin{figure*}[tb]
  \captionsetup[subfigure]{labelformat=parens, justification=centering}
  \centering
  \begin{subfigure}[b]{0.49\textwidth}
    \centering
    \includegraphics[width=1\textwidth]{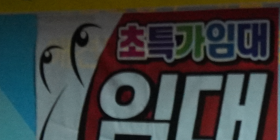}
    \caption{Reference}
  \end{subfigure}
  \hfill
  \begin{subfigure}[b]{0.49\textwidth}
      \centering
      \includegraphics[width=1\textwidth]{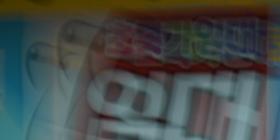}
      \caption{Blurred}
  \end{subfigure}
  \hfill
  \begin{subfigure}[b]{0.49\textwidth}
      \centering
      \includegraphics[width=1\textwidth]{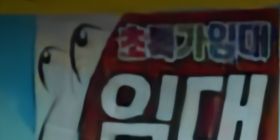}
      \caption{MPRNet~\cite{zamir2021multi}}
  \end{subfigure}
  \hfill
  \begin{subfigure}[b]{0.49\textwidth}
      \centering
      \includegraphics[width=1\textwidth]{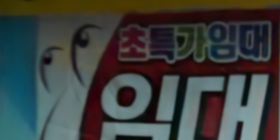}
      \caption{DeepRFT~\cite{mao2023intriguing}}
  \end{subfigure}
  \hfill
  \begin{subfigure}[b]{0.49\textwidth}
      \centering
      \includegraphics[width=1\textwidth]{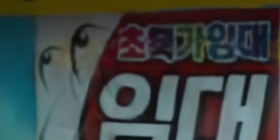}
      \caption{UFPNet~\cite{fang2023self}}
  \end{subfigure}
  \hfill
  \begin{subfigure}[b]{0.49\textwidth}
      \centering
      \includegraphics[width=1\textwidth]{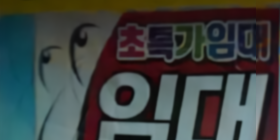}
      \caption{MISC~\cite{liu2024motion}}
  \end{subfigure}
  \hfill
  \begin{subfigure}[b]{0.49\textwidth}
      \centering
      \includegraphics[width=1\textwidth]{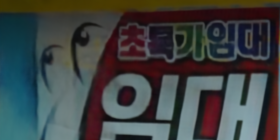}
      \caption{LoFormer~\cite{mao2024loformer}}
  \end{subfigure}
  \hfill
  \begin{subfigure}[b]{0.49\textwidth}
      \centering
      \includegraphics[width=1\textwidth]{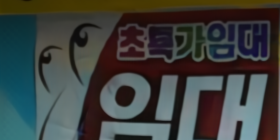}
      \caption{\bf Ours}
  \end{subfigure}
    \caption{Qualitative results of deblurring models trained and tested on RealBlur-J~\cite{rim2020real}.}\label{fig:sota_sup_rbj_3}
\end{figure*}

\begin{figure*}[tb]
  \captionsetup[subfigure]{labelformat=parens, justification=centering}
  \centering
  \begin{subfigure}[b]{0.49\textwidth}
    \centering
    \includegraphics[width=1\textwidth]{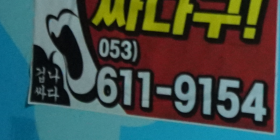}
    \caption{Reference}
  \end{subfigure}
  \hfill
  \begin{subfigure}[b]{0.49\textwidth}
      \centering
      \includegraphics[width=1\textwidth]{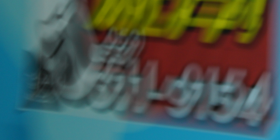}
      \caption{Blurred}
  \end{subfigure}
  \hfill
  \begin{subfigure}[b]{0.49\textwidth}
      \centering
      \includegraphics[width=1\textwidth]{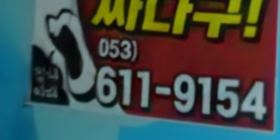}
      \caption{DeepRFT~\cite{mao2023intriguing}}
  \end{subfigure}
  \hfill
  \begin{subfigure}[b]{0.49\textwidth}
      \centering
      \includegraphics[width=1\textwidth]{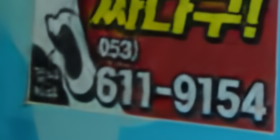}
      \caption{UFPNet~\cite{fang2023self}}
  \end{subfigure}
  \hfill
  \begin{subfigure}[b]{0.49\textwidth}
      \centering
      \includegraphics[width=1\textwidth]{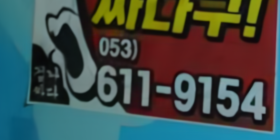}
      \caption{FFTformer~\cite{kong2023efficient}}
  \end{subfigure}
  \hfill
  \begin{subfigure}[b]{0.49\textwidth}
      \centering
      \includegraphics[width=1\textwidth]{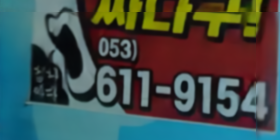}
      \caption{MISC~\cite{liu2024motion}}
  \end{subfigure}
  \hfill
  \begin{subfigure}[b]{0.49\textwidth}
      \centering
      \includegraphics[width=1\textwidth]{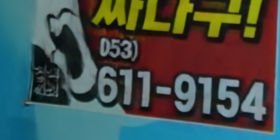}
      \caption{LoFormer~\cite{mao2024loformer}}
  \end{subfigure}
  \hfill
  \begin{subfigure}[b]{0.49\textwidth}
      \centering
      \includegraphics[width=1\textwidth]{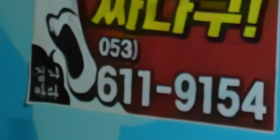}
      \caption{\bf Ours}
  \end{subfigure}
    \caption{Qualitative results of deblurring models trained and tested on RealBlur-J~\cite{rim2020real}.}\label{fig:sota_sup_rbj_4}
\end{figure*}

\begin{figure*}[tb]
  \captionsetup[subfigure]{labelformat=parens, justification=centering}
  \centering
  \begin{subfigure}[b]{0.49\textwidth}
    \centering
    \includegraphics[width=1\textwidth]{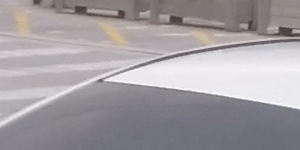}
    \caption{Reference}
  \end{subfigure}
  \hfill
  \begin{subfigure}[b]{0.49\textwidth}
      \centering
      \includegraphics[width=1\textwidth]{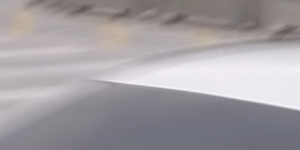}
      \caption{Blurred}
  \end{subfigure}
  \hfill
  \begin{subfigure}[b]{0.49\textwidth}
      \centering
      \includegraphics[width=1\textwidth]{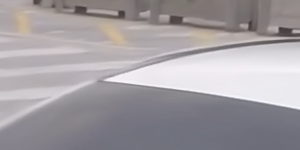}
      \caption{NAFNet~\cite{chen2022simple}}
  \end{subfigure}
  \hfill
  \begin{subfigure}[b]{0.49\textwidth}
      \centering
      \includegraphics[width=1\textwidth]{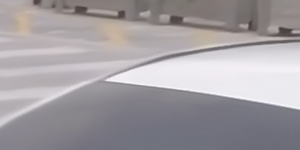}
      \caption{Restormer~\cite{zamir2022restormer}}
  \end{subfigure}
  \hfill
  \begin{subfigure}[b]{0.49\textwidth}
      \centering
      \includegraphics[width=1\textwidth]{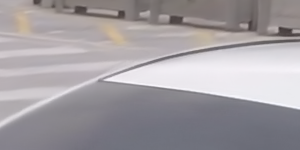}
      \caption{UFPNet~\cite{fang2023self}}
  \end{subfigure}
  \hfill
  \begin{subfigure}[b]{0.49\textwidth}
      \centering
      \includegraphics[width=1\textwidth]{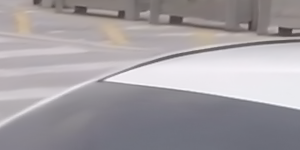}
      \caption{FFTformer~\cite{kong2023efficient}}
  \end{subfigure}
  \hfill
  \begin{subfigure}[b]{0.49\textwidth}
      \centering
      \includegraphics[width=1\textwidth]{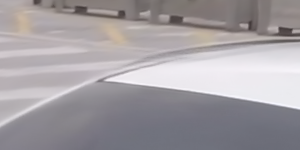}
      \caption{LoFormer~\cite{mao2024loformer}}
  \end{subfigure}
  \hfill
  \begin{subfigure}[b]{0.49\textwidth}
      \centering
      \includegraphics[width=1\textwidth]{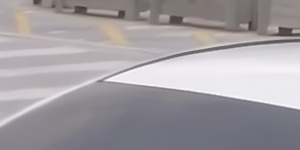}
      \caption{\bf Ours}
  \end{subfigure}
    \caption{Qualitative results of GoPro-trained deblurring models tested on GoPro~\cite{nah2017deep}. The above images show that ours restores sharper edges.}\label{fig:sota_sup_hide_1}
\end{figure*}

\begin{figure*}[tb]
  \captionsetup[subfigure]{labelformat=parens, justification=centering}
  \centering
  \begin{subfigure}[b]{0.49\textwidth}
    \centering
    \includegraphics[width=1\textwidth]{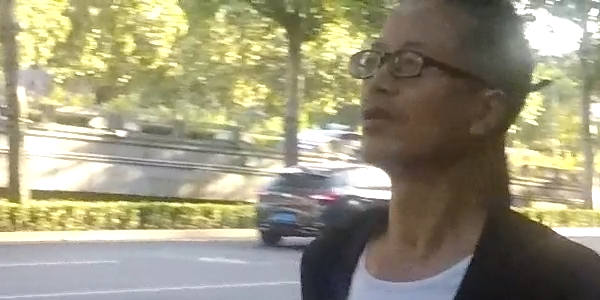}
    \caption{Reference}
  \end{subfigure}
  \hfill
  \begin{subfigure}[b]{0.49\textwidth}
      \centering
      \includegraphics[width=1\textwidth]{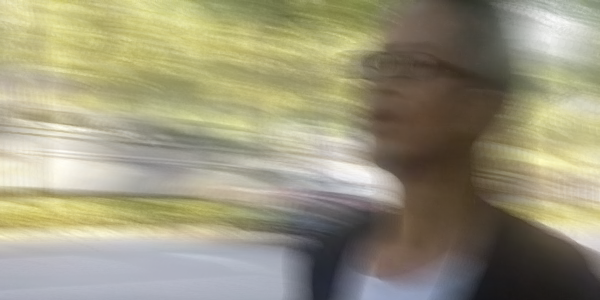}
      \caption{Blurred}
  \end{subfigure}
  \hfill
  \begin{subfigure}[b]{0.49\textwidth}
      \centering
      \includegraphics[width=1\textwidth]{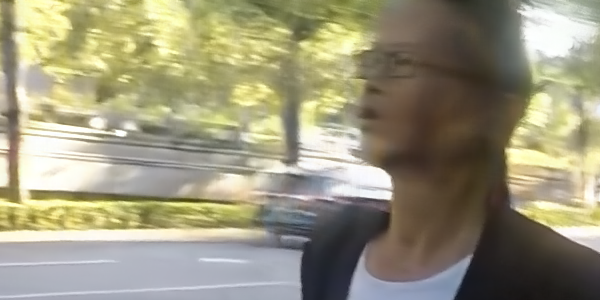}
      \caption{NAFNet~\cite{chen2022simple}}
  \end{subfigure}
  \hfill
  \begin{subfigure}[b]{0.49\textwidth}
      \centering
      \includegraphics[width=1\textwidth]{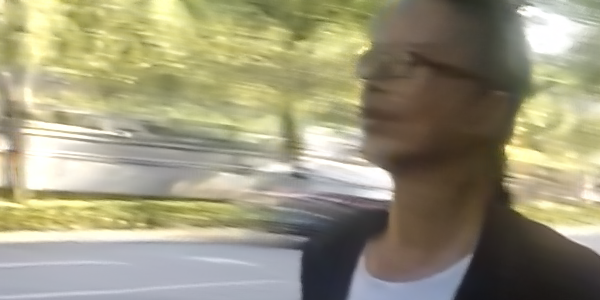}
      \caption{Restormer~\cite{zamir2022restormer}}
  \end{subfigure}
  \hfill
  \begin{subfigure}[b]{0.49\textwidth}
      \centering
      \includegraphics[width=1\textwidth]{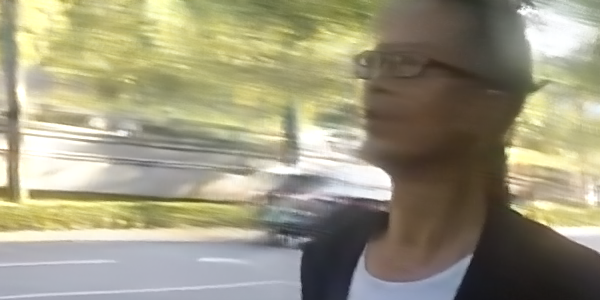}
      \caption{UFPNet~\cite{fang2023self}}
  \end{subfigure}
  \hfill
  \begin{subfigure}[b]{0.49\textwidth}
      \centering
      \includegraphics[width=1\textwidth]{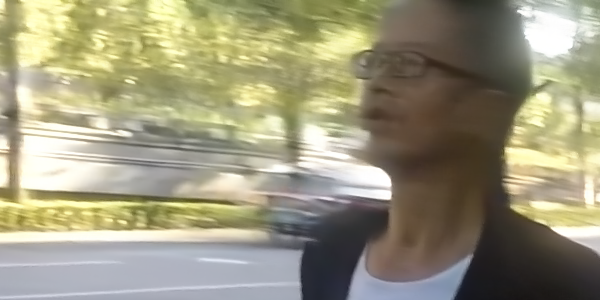}
      \caption{FFTformer~\cite{kong2023efficient}}
  \end{subfigure}
  \hfill
  \begin{subfigure}[b]{0.49\textwidth}
      \centering
      \includegraphics[width=1\textwidth]{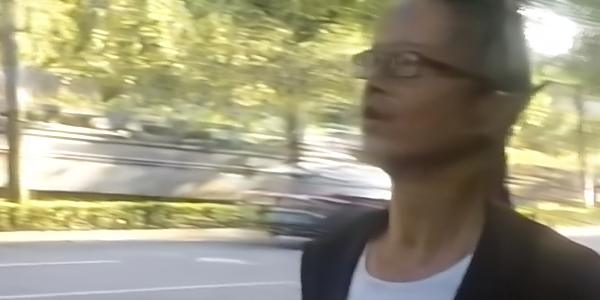}
      \caption{LoFormer~\cite{mao2024loformer}}
  \end{subfigure}
  \hfill
  \begin{subfigure}[b]{0.49\textwidth}
      \centering
      \includegraphics[width=1\textwidth]{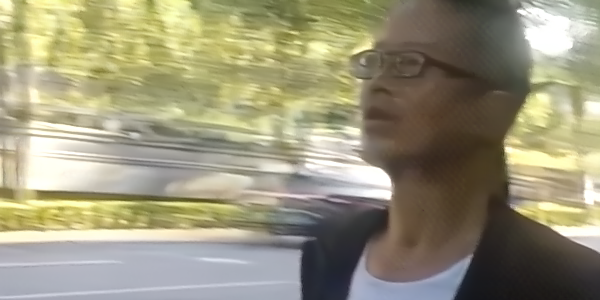}
      \caption{\bf Ours}
  \end{subfigure}
    \caption{Qualitative results of GoPro-trained deblurring models tested on HIDE~\cite{shen2019human}. The above images show that ours restores clearer facial details.}\label{fig:sota_sup_gopro_1}
\end{figure*}

\begin{figure*}[tb]
  \captionsetup[subfigure]{labelformat=parens, justification=centering}
  \centering
  \begin{subfigure}[b]{0.49\textwidth}
    \centering
    \includegraphics[width=1\textwidth]{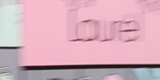}
    \caption{Blurred}
  \end{subfigure}
  \hfill
  \begin{subfigure}[b]{0.49\textwidth}
      \centering
      \includegraphics[width=1\textwidth]{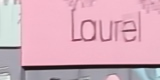}
      \caption{MPRNet~\cite{zamir2021multi}}
  \end{subfigure}
  \hfill
  \begin{subfigure}[b]{0.49\textwidth}
      \centering
      \includegraphics[width=1\textwidth]{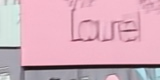}
      \caption{NAFNet~\cite{chen2022simple}}
  \end{subfigure}
  \hfill
  \begin{subfigure}[b]{0.49\textwidth}
      \centering
      \includegraphics[width=1\textwidth]{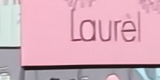}
      \caption{Restormer~\cite{zamir2022restormer}}
  \end{subfigure}
  \hfill
  \begin{subfigure}[b]{0.49\textwidth}
      \centering
      \includegraphics[width=1\textwidth]{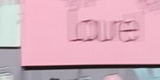}
      \caption{FFTformer~\cite{kong2023efficient}}
  \end{subfigure}
  \hfill
  \begin{subfigure}[b]{0.49\textwidth}
      \centering
      \includegraphics[width=1\textwidth]{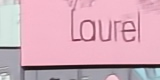}
      \caption{UFPNet~\cite{fang2023self}}
  \end{subfigure}
  \hfill
  \begin{subfigure}[b]{0.49\textwidth}
      \centering
      \includegraphics[width=1\textwidth]{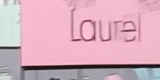}
      \caption{LoFormer~\cite{mao2024loformer}}
  \end{subfigure}
  \hfill
  \begin{subfigure}[b]{0.49\textwidth}
      \centering
      \includegraphics[width=1\textwidth]{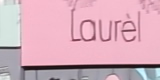}
      \caption{\bf Ours}
  \end{subfigure}
    \caption{Qualitative results of GoPro-trained deblurring models tested on RWBI~\cite{zhang2020deblurring}.}\label{fig:sota_sup_rwbi_1}
\end{figure*}

In this section, we begin with visual results of deblurring models trained and tested on the RealBlur-R dataset~\cite{rim2020real}, which have not been included in the main paper due to space limits. The low-light images may not be clearly visible in this material. Then, an extensive collection of deblurred images of different networks are appended to this material, which demonstrate the impressive visuals of XYScanNet on mainstream datasets~\cite{nah2017deep,shen2019human,zhang2020deblurring,rim2020real}. Please refer to the current and next pages for additional visual results, as shown in~\cref{fig:sota_sup_rbr_1}-\cref{fig:sota_sup_rwbi_1}.


\end{document}